\documentclass[10pt,twocolumn,letterpaper]{article}

\usepackage[pagenumbers]{cvpr} % To force page numbers, e.g. for an arXiv version
\usepackage{multirow} 
\usepackage{hyperref}       % hyperlinks
\usepackage{url}            % simple URL typesetting
\usepackage{booktabs}       % professional-quality tables
\usepackage{amsfonts}       % blackboard math symbols
\usepackage{nicefrac}       % compact symbols for 1/2, etc.
\usepackage{microtype}      % microtypography
\usepackage{xcolor}         % colors

\usepackage{graphicx}
\usepackage{booktabs}

\usepackage{multirow}
\usepackage{pifont}
\usepackage{threeparttable}
\usepackage{amsmath}
\usepackage{enumitem}

\usepackage{algorithm}
\usepackage{algpseudocode}

\PassOptionsToPackage{dvipsnames}{xcolor}

\usepackage{graphicx}
\usepackage{booktabs}

\usepackage{multirow}
\usepackage{pifont}
\usepackage{threeparttable}
\usepackage{amsmath}
\usepackage{enumitem}

\usepackage{algorithm}
\usepackage{algpseudocode}

\definecolor{darkred}{HTML}{C00000}
\definecolor{darkgreen}{HTML}{385723}
\definecolor{darkblue}{HTML}{1F4E79}
\definecolor{lightblue}{HTML}{DEEBF7}

\definecolor{cvprblue}{rgb}{0.21,0.49,0.74}

\usepackage[accsupp]{axessibility}

\title{CoD: A Diffusion Foundation Model for Image Compression}

\author{Zhaoyang Jia$^1$\footnotemark[1] \quad\! Zihan Zheng$^1$\footnotemark[1] \quad\! Naifu Xue$^2$\footnotemark[1] \quad\! Jiahao Li$^3\quad\!$ Bin Li$^3$ \\
Zongyu Guo$^3\quad\!$Xiaoyi Zhang$^3 \quad\!$ Houqiang Li$^1 \quad\!$ Yan Lu$^3$
\vspace{1mm}\\
$^1$ University of Science and Technology of China $\ \ ^2$ Communication University of China\\
$^3$ Microsoft Research Asia\\
{\tt\small \{jzy\_ustc, zzh2003\}@mail.ustc.edu.cn, lihq@ustc.edu.cn}, 
{\tt\small \{aaronxuenf\}@cuc.edu.cn}
\\
{\tt\small \{libin, li.jiahao, zongyuguo, xiaoyizhang, yanlu\}@microsoft.com}
}

\begin{document}

\newlist{myitemize}{itemize}{1}
\setlist[myitemize,1]{label=\textbullet,leftmargin=5.5mm}

% \maketitle

\twocolumn[{%
\renewcommand\twocolumn[1][]{#1}%
\maketitle
\centering
\includegraphics[width=\linewidth]{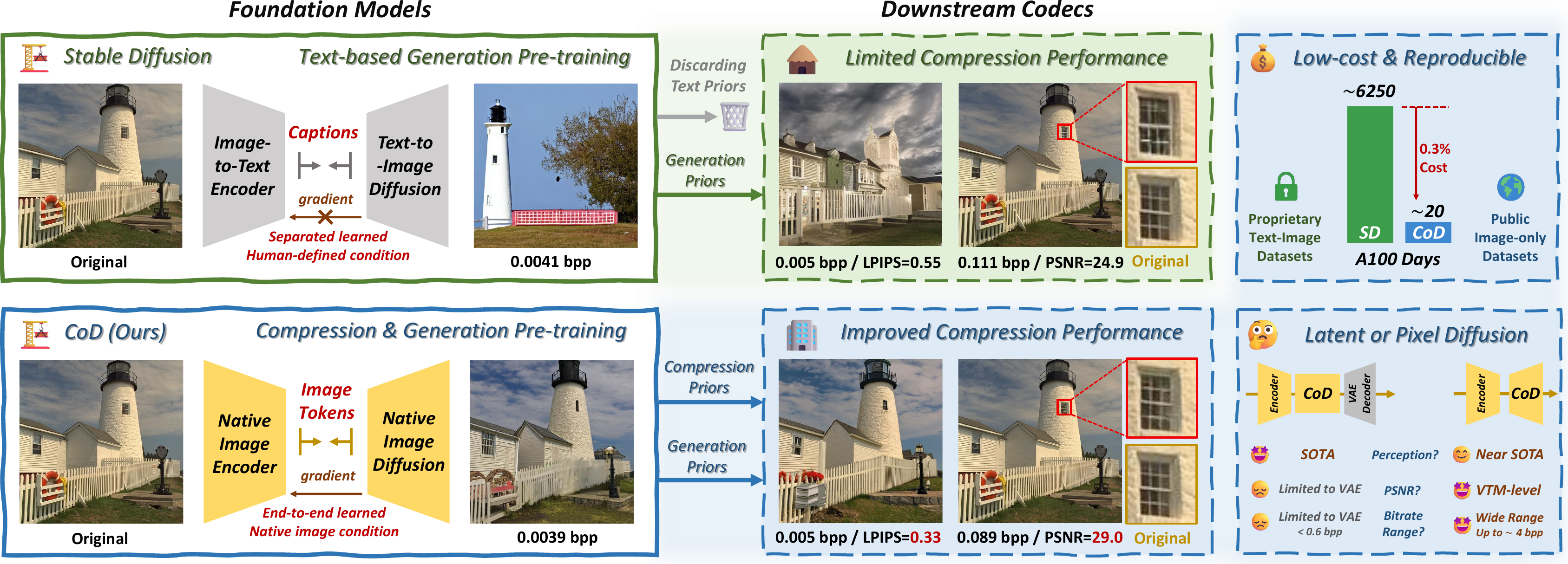}
\vspace{-6mm}
\captionof{figure}{Overview of \textbf{Co}mpression-oriented \textbf{D}iffusion (CoD) foundation models, which are trained from scratch to jointly optimize compression and generation. Rather than a fixed codec, CoD serves as a foundational model for downstream diffusion-based codecs such as DiffC~\cite{diffc_original}, substantially enhancing their performance by replacing Stable Diffusion.
}
\label{fig:Foundation}
\vspace{4mm}
}]

{
    \renewcommand{\thefootnote}{\fnsymbol{footnote}}
    \footnotetext[1]{This work was done when Zhaoyang Jia, Zihan Zheng and Naifu Xue were full-time interns at Microsoft Research Asia.}
}

\begin{abstract}
\vspace{-2mm}
Existing diffusion codecs typically build on text-to-image diffusion foundation models like Stable Diffusion.
However, text conditioning is suboptimal from a compression perspective, hindering the potential of downstream diffusion codecs, particularly at ultra-low bitrates.
To address it, we introduce \textbf{CoD}, the first \textbf{Co}mpression-oriented \textbf{D}iffusion foundation model, trained from scratch to enable end-to-end optimization of both compression and generation. CoD is not a fixed codec but a general foundation model designed for various diffusion-based codecs.
It offers several advantages: \textbf{High compression efficiency}, replacing Stable Diffusion with CoD in downstream codecs like DiffC achieves SOTA results, especially at ultra-low bitrates (e.g., 0.0039 bpp); \textbf{Low-cost and reproducible training}, 300$\times$ faster training than Stable Diffusion ($\sim$ 20 vs. $\sim$ 6,250 A100 GPU days) on entirely open image-only datasets; \textbf{Providing new insights}, e.g., We find pixel-space diffusion can achieve VTM-level PSNR with high perceptual quality and can outperform GAN-based codecs using fewer parameters.
We hope CoD lays the foundation for future diffusion codec research.
% Codes will be released.
Codes are released at \url{https://github.com/microsoft/GenCodec/tree/main/CoD}.
\end{abstract}    
\section{Introduction}
\label{sec:introduction}

Diffusion image compression~\cite{cdc, perco, perco_sd, diffeic, diffc, ddcm, oscar} exploits the strong generative priors of denoising diffusion models~\cite{diffusion, score_matching, rf} to achieve realistic image reconstruction. To inherit such generative priors, recent diffusion codecs typically adopt large-scale pretrained text-to-image diffusion models such as Stable Diffusion~\cite{SD}, demonstrating impressive performance in generative image compression.

However, when using text-to-image diffusion models in compression, the role of text conditions remains unclear. Although early works such as Text+Sketch~\cite{text+sketch} and PerCo~\cite{perco} claim image captions can serve as a high-level semantic signal to facilitate ultra-low bitrate compression, later studies~\cite{onedc, oscar} demonstrate that fine-tuning Stable Diffusion to discard text priors can further improve compression efficiency. Additional evidence comes from the zero-shot diffusion compression framework DiffC~\cite{diffc}, which theoretically measures the “compression capability” of a denoising diffusion model. When applied to Stable Diffusion, DiffC reveals that text conditions are detrimental to compression performance, especially at low bitrates. These observations suggest that text-conditioned diffusion models are not naturally suitable for compression, and that \textcolor{darkred}{\textbf{Stable Diffusion is not the ideal foundation model for diffusion codecs}}.

To understand this phenomenon, we analyze the underlying mechanism. In a standard compression formulation,
\begin{equation}
    y = \mathrm{Encode}(x, \Theta), \quad 
    \hat{x} = \mathrm{Decode}(y, \Phi)
\end{equation}
where $x$, $y$, and $\hat{x}$ denote the original image, compressed representation, and reconstruction, respectively, and $\Theta$ and $\Phi$ represent the trained parameters of the encoder and decoder. If we set $\Theta$ as an image captioner like BLIP-2~\cite{blip} and $\Phi$ as Stable Diffusion, we note that \textit{text-conditioned diffusion can be formulated in the compression formulation}, as illustrated in the top left of Figure~\ref{fig:Foundation}. From the perspective of compression, text conditions introduces two limitations. First, human-generated text is less efficient in describing fine-grained spatial and texture details of natural images. Second, the discrete vocabularies in text is non-differentiable to prevent joint optimization of $\Theta$ and $\Phi$, thus making rate–distortion optimization inapplicable.

Recognizing the limitations of text conditions, a natural next step is to \textcolor{darkred}{\textbf{train a diffusion foundation model explicitly oriented toward compression}}. Neural codecs can learn compact, native image tokens as conditions for diffusion models, and enable end-to-end training with the diffusion model to determine the optimal image representations for coding. By carefully designing and leveraging advanced diffusion training algorithms, we introduce the first \textbf{Co}mpression-oriented \textbf{D}iffusion foundation model, \textbf{CoD}. The architecture of CoD is simple yet effective: a native image encoder $\Theta$ to compress images into tokens, followed by an information bottleneck, and a diffusion model $\Phi$ to decode pixels conditioned on these image tokens. The bottleneck is designed to enforce ultra-low bitrates (e.g., 0.0039 bpp), transmitting only essential semantic information while allowing the diffusion model the flexibility to generate realistic details. As illustrated in the bottom left of Figure~\ref{fig:Foundation}, CoD achieves significantly higher reconstruction fidelity compared to text-to-image foundation models. 

It is worth noting that CoD is not a fixed codec but a general foundation model designed for diverse diffusion-based compression algorithms. In practice, it can replace Stable Diffusion in existing downstream diffusion codecs to improve performance. Unlike Stable Diffusion that only provides generation priors, CoD additionally learns compression priors to substantially boost downstream performance, particularly at low bitrates.
CoD offers several advantages:

\begin{figure*}[t]
  \centering
    \includegraphics[width=\linewidth]{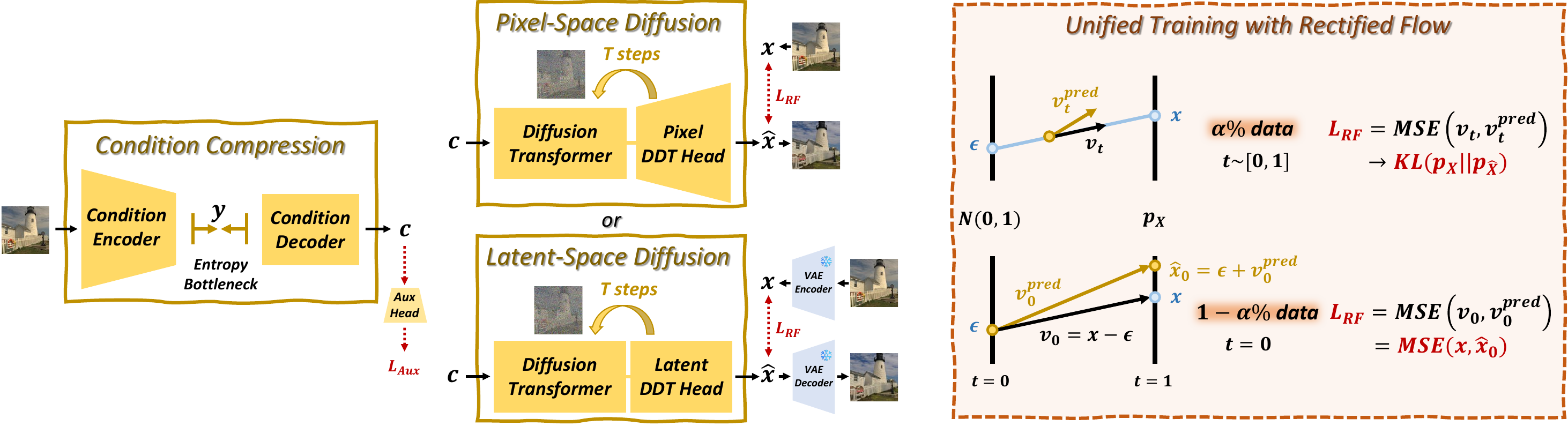}
    \vspace{-3mm}
    \caption{Framework overview of CoD in pixel and latent spaces. CoD consists of a condition encoder, an entropy bottleneck, a condition decoder and a diffusion model which is decoupled to DiT backbone and DDT head~\cite{ddt}. CoD is trained with rectified flow~\cite{rf}, where $1-\alpha\%$ samples are trained at $t=0$ to jointly optimize distortion and perception. } 
  \label{fig:Framework}
  \vspace{-2mm}
\end{figure*}

\textbf{Low Training Cost}.
Compared to Stable Diffusion, CoD is far more efficient to train. Stable Diffusion v1.5 requires about 6,250 A100 GPU days~\cite{pixart_alpha}, whereas CoD only takes around 20 A100 GPU days (0.3\% cost). Unlike text-to-image diffusion that relies on text-image training pairs, CoD is fully self-supervised and only requires image datasets, making it easier to scale. Furthermore, while Stable Diffusion uses proprietary datasets, CoD is trained entirely on open datasets such as ImageNet~\cite{imagenet}, OpenImages~\cite{openimage}, and SA-1B~\cite{sa1b}, making it easy to reproduce and unleashing the potential for future exploration across different diffusion codec architectures and training schemes.

\textbf{Providing New Insights.}
We conduct an in-depth analysis of CoD to gain insights into diffusion-based compression. We facilitate scaling-law study of CoD to reveal how model size affects compression performance. We further show that diffusion codecs can surpass GAN-based codecs even with significantly fewer parameters. In addition, we present a comprehensive comparison between pixel-space and latent-space diffusion. While latent diffusion offers superior perceptual quality at low bitrates, its PSNR and bitrate range are constrained by the VAE (e.g., up to $\sim$26 dB PSNR at 0.6 bpp). In contrast, pixel diffusion delivers both high perceptual quality and high PSNR across a wide bitrate range (e.g., reaching near-lossless $\sim$47 dB PSNR at 4 bpp).

\textbf{High Compression Performance}.
As a compression-oriented foundation model, CoD enables high-performance downstream codecs. In this paper, we evaluate it using the state-of-the-art zero-shot diffusion compression method, DiffC~\cite{diffc_original, diffc}. Compared to Stable Diffusion, DiffC with CoD achieves significantly better performance, especially at low bitrates. With pixel-space CoD, it even matches VTM~\cite{vvc} in PSNR while outperforming previous pixel-space perceptual codecs such as MS-ILLM~\cite{illm} in FID. The qualitative results in middle column of Figure~\ref{fig:Foundation} demonstrate that pixel-space CoD achieves much higher reconstruction fidelity, highlighting the potential of pixel diffusion to approach the theoretical rate–distortion–perception trade-off.

The contributions of this work can be summarized as:
\begin{itemize}
\item We introduce the first compression-oriented diffusion foundation model, CoD, designed to replace Stable Diffusion in diffusion codecs.
\item CoD enables low-cost, reproducible training, supporting exploration of new network architectures.
\item CoD delivers high compression performance, especially on ultra-low bitrates.
\item CoD pushes the boundaries of diffusion codecs, demonstrating the potential of pixel-space diffusion as a  pathway toward the optimal rate–distortion–perception trade-off.
\end{itemize}

\section{Compression-oriented Diffusion}
\label{sec:method}

In this section, we introduce the implementation details of the proposed compression-oriented diffusion models, CoD.

\subsection{Pipeline}
\label{sec:pipeline}

We implement CoD in both pixel and latent spaces. As illustrated in Figure~\ref{fig:Framework}, given an input data $x$, CoD performs:

\noindent\textbf{Condition Encoding}. An encoder firstly compresses $x$ into compact representations. Following~\cite{vqgan}, we stack residual blocks~\cite{resnet} and attention layers~\cite{transformer} to construct the encoder. The output is at $1/32$ resolution of the original images.

\noindent\textbf{Entropy bottleneck}. 
It is designed to maintain an ultra-low bitrate for $y$, encouraging the diffusion model to learn strong generative capabilities. This can be implemented as scalar quantization with entropy models~\cite{hyperprior}, vector quantization~\cite{vqgan}, or finite scalar quantization~\cite{fsq}. For simplicity, we use vector quantization with a codebook size of $N=2^4=16$, corresponding to $4\text{ bits} / (32\times32) = 0.0039\text{ bpp}$.

\noindent\textbf{Condition Decoding}. The condition decoder reconstructs intermediate compression conditions $c$ from the image tokens $y$. The decoder adopts similar structures as the encoder. The condition $c$ is at a $1/16$ resolution of the original image.

\noindent\textbf{Diffusion-based reconstruction}. The diffusion module denoises a randomly sampled Gaussian noise over $T$ steps under the guidance of condition $c$ to reconstruct $\hat{x}$. Following DDT~\cite{ddt}, we decouple DiT backbone and DDT head to enhance capacity. DDT performs denoising at $1/16$ resolution, where the condition $c$ is concatenated with the noised input along the channel dimension at each step for guidance.

In the latent space, since VAE latents are already at $1/8$ resolution, we apply a $2\times2$ patch embedding to downsample them to $1/16$, and upsample the output of DDT head back to $1/8$ before passing to the VAE decoder. In the pixel space, we directly use a $16\times16$ patch embedding to map the noised images to $1/16$ resolution, and adopt the pixel DDT head from~\cite{pixnerd}, where each feature predicts a neural field that reconstructs a corresponding $16\times16$ patch. Since the DiT backbone runs at $1/16$ resolution of original images, \textbf{pixel-space CoD has a similar computation complexity as in latent space}. Details are in the appendix.

\begin{figure}[t]
  \centering
    \includegraphics[width=\linewidth]{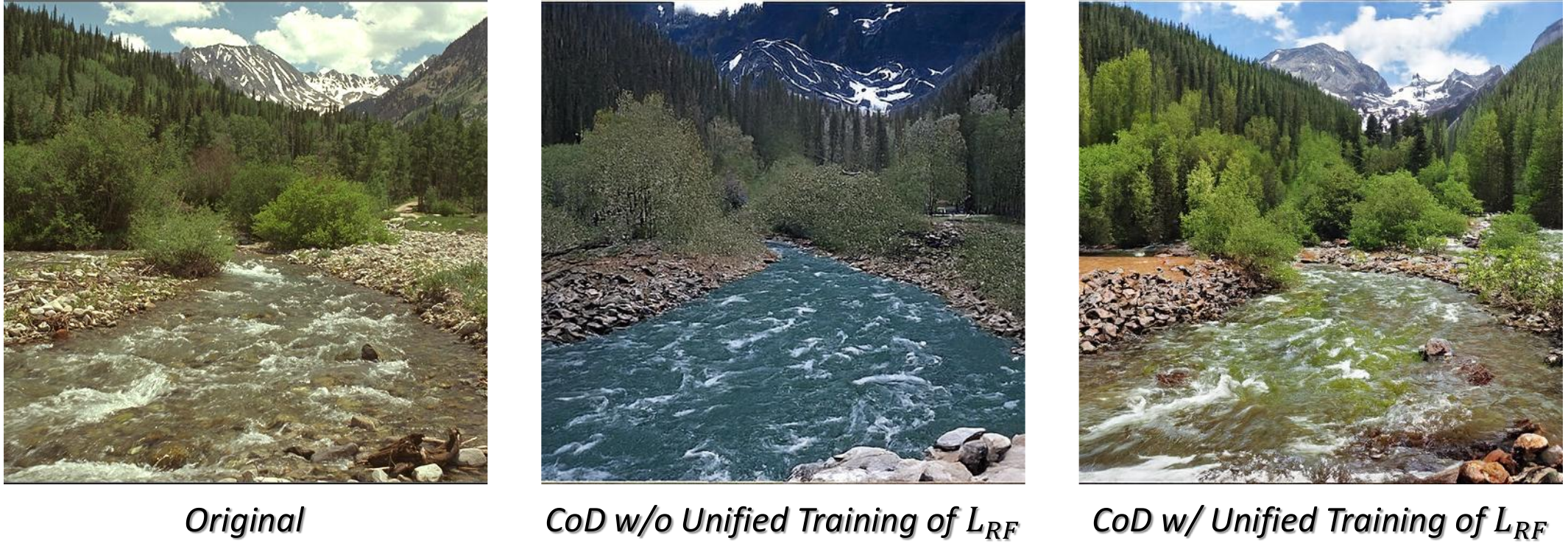}
    \vspace{-4mm}
    \caption{Effects of unified training with rectified flow.} 
    \vspace{-2mm}
  \label{fig:Compare_Unified_Traininig}
\end{figure}

\begin{figure*}[t]
  \centering
    \includegraphics[width=\linewidth]{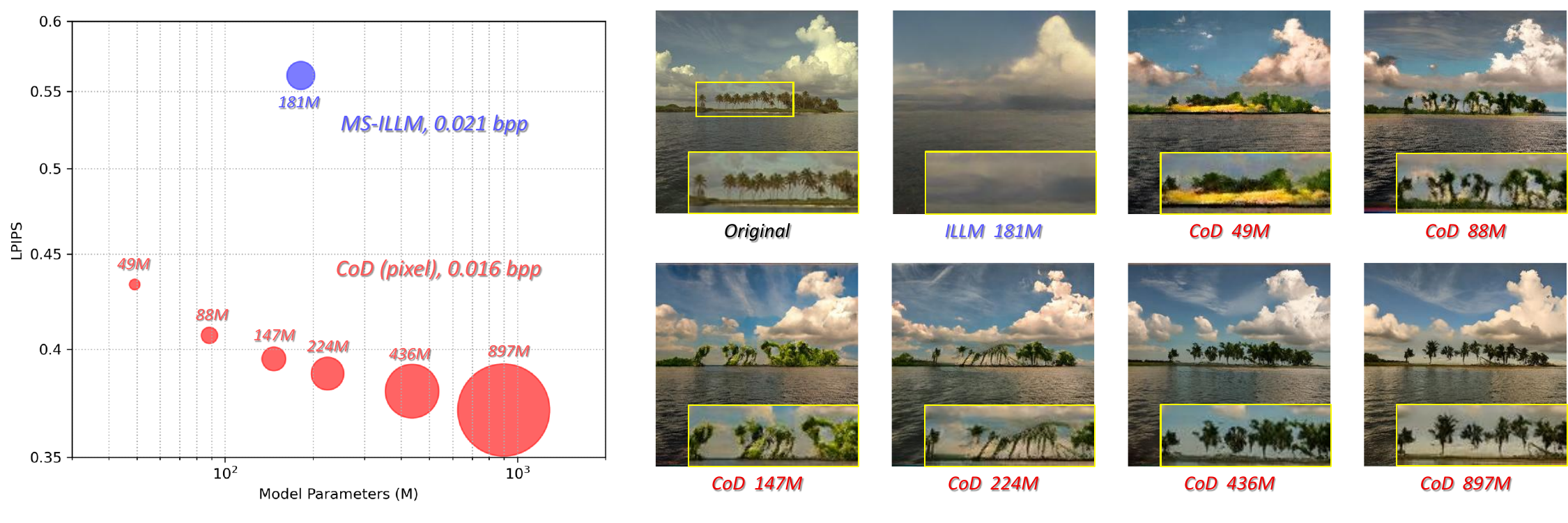}
    \vspace{-7mm}
    \caption{Scaling law analysis on Kodak at $256\times256$. All CoD models are at 0.016 bpp while MS-ILLM is at 0.021 bpp.} 
    \vspace{-2mm}
  \label{fig:Scaling_Law}
\end{figure*}

\subsection{Diffusion Training}

\label{sec:diffusion_training}

\noindent\textbf{Preliminary}.
Denoising diffusion models~\cite{diffc} model the data distribution by progressively perturbing a clean sample $x$ with Gaussian noise $\varepsilon$, transforming $p(x)$ into a standard Gaussian distribution. At timestep $t$, it is formulated as, 
\begin{equation}
    x_t = \alpha_t x + \sigma_t \varepsilon, \quad 
    \varepsilon \sim \mathcal{N}(0, I)
\end{equation}
where $\alpha_t$ and $\sigma_t$ define the noise schedule. The process can be expressed as a stochastic differential equation (SDE),
\begin{equation}
    \mathrm{d}x_t = f(t)\,x_t\,\mathrm{d}t + g(t)\,\mathrm{d}\omega_t
\end{equation}
with drift $f(t)$ and diffusion $g(t)$ determined by $\alpha_t$ and $\sigma_t$.
Score-based models~\cite{score_matching} learn the score function of $p(x)$ to solve the reverse SDE and generate samples.

Rectified Flow (RF)~\cite{rf} reinterprets diffusion as a deterministic transformation. It removes the stochastic term from the SDE and adopts a linear transition schedule:
\begin{equation}
x_t = t \cdot x + (1-t) \cdot \varepsilon.
\end{equation}
So the trajectory follows a simple ordinary differential equation (ODE) formulated as:
\begin{equation}
    \mathrm{d}x_t = v_t\,\mathrm{d}t, 
    \quad \text{where } v_t = x - \varepsilon
\end{equation}
Instead of estimating the score, RF directly learns the velocity field $v_t$, providing more stable training, faster inference, and a clearer geometric view of generation.

\noindent\textbf{Unified Training with Rectified Flow}.
CoD predicts the velocity $v_t$ so it can be trained with the rectified flow loss,
\begin{equation}
    \mathcal{L}_{\mathrm{RF}} = \mathrm{MSE}(v_t, v_t^{\mathrm{pred}})
\end{equation}
where $v_t^{\mathrm{pred}}$ denotes the predicted velocity field. Theoretically, optimizing $\mathcal{L}_{\mathrm{RF}}$ reduces the Kullback–Leibler (KL) divergence between the reconstructed and original distributions~\cite{gentiloni2024theoretical}, which can reflect the perception term in rate-distortion-perception (R-D-P) theory~\cite{rdp}. Hence, optimizing $\mathcal{L}_{\mathrm{RF}}$ naturally promotes perceptual quality.

However, we find that rectified flow loss mainly guarantee the reconstruction consistency of the structure instead of color information (see Figure~\ref{fig:Compare_Unified_Traininig}). From a compression viewpoint, this is consistent with R–D–P theory: standard diffusion training neglects the distortion term, leading to limited reconstruction fidelity.

To overcome this, we introduce a unified formulation that integrates a distortion term into the rectified flow loss. Since the one-step estimate at $t=0$ is $\hat{x}_{0}=\varepsilon+v_0^{\mathrm{pred}}$, the rectified flow loss can be rewritten given $v_{0}=x-\varepsilon$:
\begin{equation}
    \mathcal{L}_{\mathrm{RF}} = \mathrm{MSE}(v_0, v_0^{\mathrm{pred}}) = \mathrm{MSE}(x, \hat{x}_0)
\end{equation}
Thus, optimizing $\mathcal{L}_{\mathrm{RF}}$ at $t=0$ directly minimizes one-step reconstruction distortion to incorporate a distortion term during training. As illustrated in Figure~\ref{fig:Framework}, during training we randomly select $\alpha\%$ samples with $t\in[0, 1]$ and the rest with $t=0$ to enable the unified training.

\noindent\textbf{Discussion}.
Unified training is designed for continuous flows, where the probability of sampling a fully noised state $t=0$ approaches zero. We observe that color reconstruction in CoD is largely determined by its first denoising step; thus, the absence of $t=0$ optimization often leads to noticeable color shifts. In contrast, DDPM-based codecs such as PerCo~\cite{perco} employ discrete timesteps uniformly sampled as $t_{\mathrm{DDPM}}\in{1,\cdots,1000}$, inherently including optimizing of fully noised state. This can be viewed as a special case of our formulation with $1-\alpha\%=1/1000$.

\begin{figure*}[t]
  \centering
    \includegraphics[width=\linewidth]{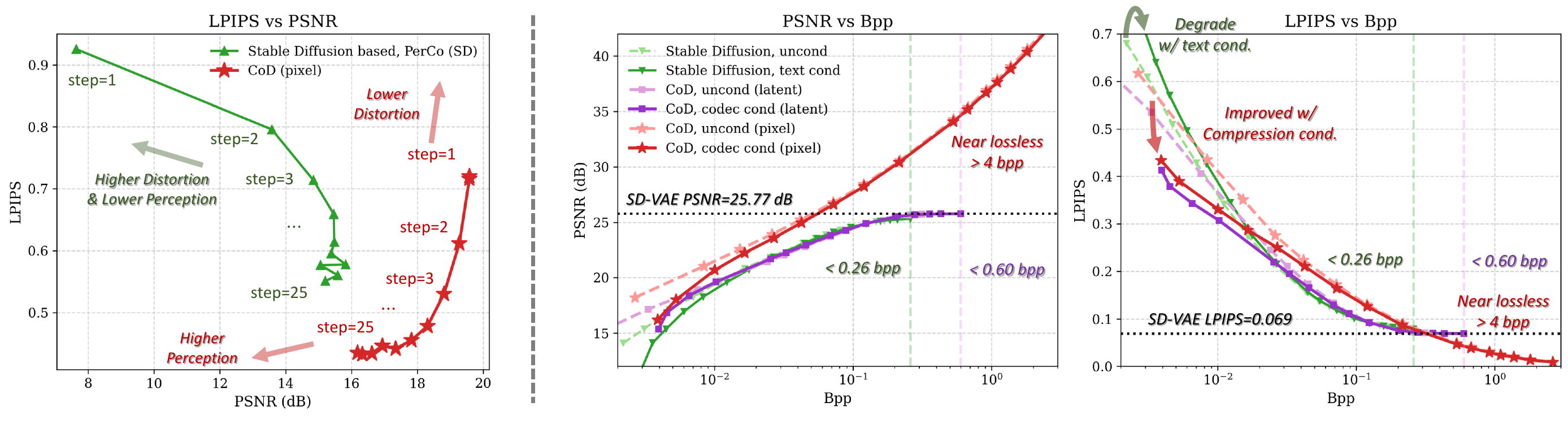}
    \vspace{-4mm}
    \caption{Comparison of CoD and Stable Diffusion on Kodak at $512\times512$ resolution. (left) Pixel-space CoD enables zero-shot distortion-perception controlling by adjusting the sampling steps. CoD is at 0.0039 bpp and PerCo is at 0.0036 bpp. (right) Text conditions harms performance of zero-shot algorithm DiffC on Stable Diffusion, while CoD condition boosts LPIPS at low-bitrate. In addition, pixel-space CoD is not limited by the SD-VAE thus demonstrating wider bitrates, higher PSNR and higher potential in perceptual quality.} 
    \vspace{-2mm}
  \label{fig:Compare_CoD_SD}
\end{figure*}

\subsection{Implementation Details}

\noindent\textbf{Datasets}.
Foundation models require large-scale training to achieve optimal performance. Unlike text-to-image generation that depends on text-image data pairs, CoD is fully self-supervised and requires only image datasets, making large-scale scaling more accessible. In this paper, we use ImageNet-21K~\cite{imagenet}, OpenImage~\cite{openimage} and SA-1B~\cite{sa1b}, including a total of 22M images. All datasets are publicly available, facilitating reproducibility for future research.

\noindent\textbf{Loss Function}. For the information bottleneck, we adopt the codebook commitment loss $\mathcal{L}_{\mathrm{C}}$~\cite{vqgan}. Diffusion training employs the rectified flow loss $\mathcal{L}_{\mathrm{RF}}$ and the representation alignment loss $\mathcal{L}_{\mathrm{REPA}}$ with DINOv2~\cite{dino_v2}. Additionally, an auxiliary head (Figure~\ref{fig:Framework}) reconstructs both original pixels and DINOv2 features from the condition $c$ to enhance condition learning. Further analysis of this auxiliary loss $\mathcal{L}_{\mathrm{aux}}$ is in the appendix. The overall objective is
\begin{equation}
\mathcal{L} = \mathcal{L}_{\mathrm{RF}} 
+ \lambda \,\cdot\, \mathcal{L}_{\mathrm{REPA}} 
+ \beta \,\cdot\, \mathcal{L}_{\mathrm{C}} 
+ \gamma \,\cdot\, \mathcal{L}_{\mathrm{aux}}
\end{equation}
where $\lambda=0.5$, $\beta=0.25$ and $\gamma=1.0$. 

\noindent\textbf{Training Details}. We incorporate several advanced techniques for diffusion training. Timesteps are sampled following a log-normal distribution, which is further enhanced by the unified training strategy introduced in Section~\ref{sec:diffusion_training}. CoD is progressively trained from low to high resolution: first at $256\times256$ for 400k steps with a batch size of 128, and then fine-tuned at $512\times512$ for 150k steps with a batch size of 64. The encoder downsamples images to $1/16$ at $256\times256$ resolution (0.0156 bpp) and to $1/32$ at $512\times512$ (0.0039 bpp), ensuring that the total number of patches and overall bit cost remain consistent.

\noindent\textbf{Training Costs}. CoD is trained on 4 NVIDIA A100 GPUs for approximately 5 days, totaling around 20 A100 GPU days. This represents only about $0.3\%$ of the training cost of Stable Diffusion v1.5~\cite{SD}, primarily because: (1) CoD learns the most informative compression conditions end-to-end, rather than relying on high-dimensional, human-defined text embeddings; (2) the unified rectified flow training stabilizes and accelerates convergence; (3) advanced optimization strategies such as REPA~\cite{repa} further improve efficiency.

\noindent\textbf{Diffusion Sampling}.  For ODE-based sampling, we employ the Adam-like-2nd solver~\cite{pixnerd} with 25 steps. We use classifier-free guidance~\cite{cfg} with scales of 3.0 for pixel space and 1.25 for latent space to enhance conditional generation.
\section{Analysis on Foundation Model}
\label{sec:eval}

In this section, we conduct an in-depth analysis of CoD to investigate its performance and underlying characteristics. More analysis are in the appendix.

\begin{figure}[t]
  \centering
    \includegraphics[width=\linewidth]{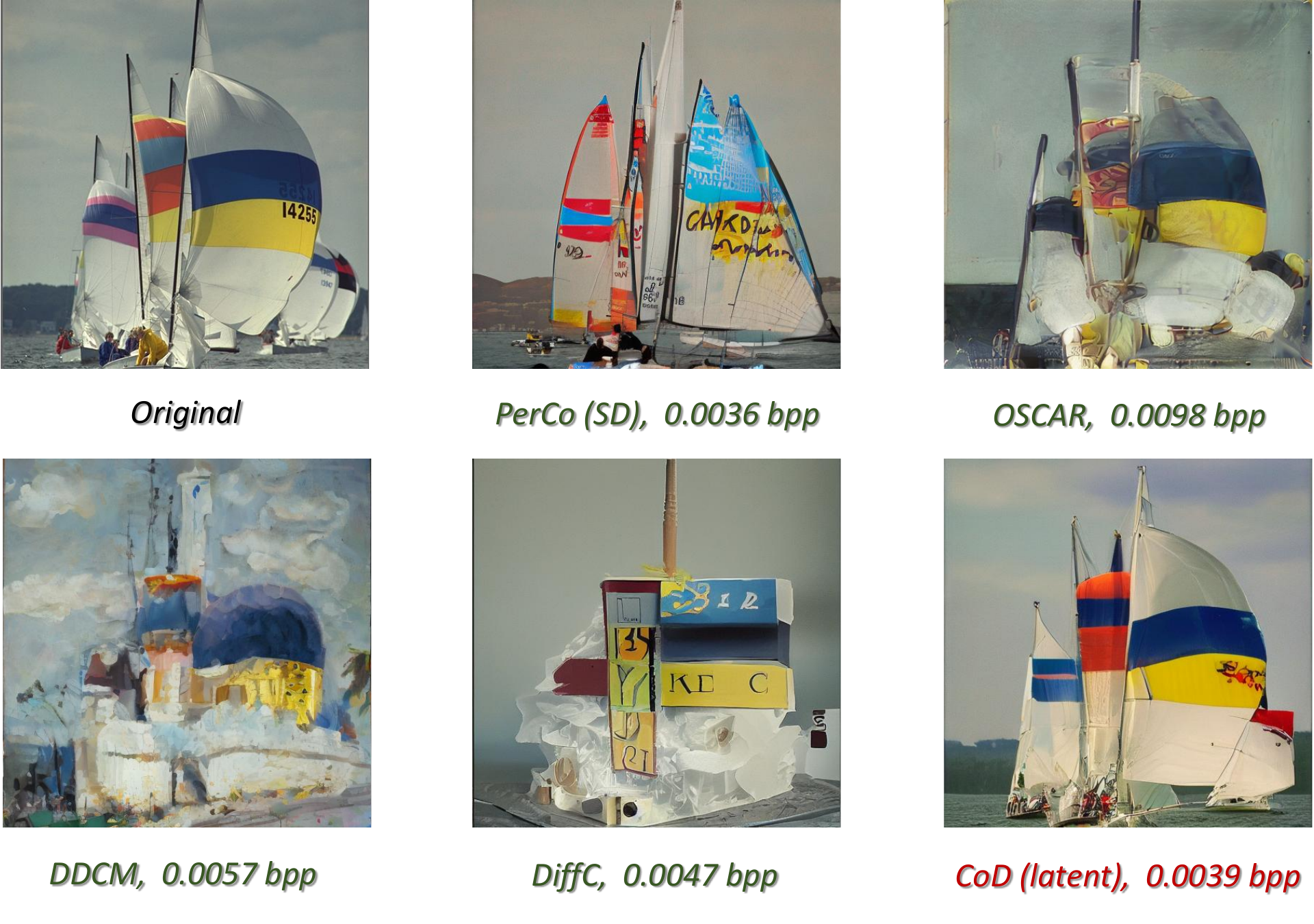}
    \vspace{-7mm}
    \caption{Visual comparison between \textcolor{darkgreen}{Stable-Diffusion-based codecs} and \textcolor{darkred}{latent-space CoD} under ultra-low bitrates.} 
    \vspace{-3mm}
  \label{fig:Eval_Fondation_Model}
\end{figure}

\subsection{Evaluation}
\label{sec:eval_method}

CoD naturally operates as a codec at 0.0039 bpp. We evaluate its compression performance on the Kodak~\cite{kodak} dataset at $512\times512$ resolution. To ensure fair comparison among foundation models, we employ the same SD-VAE and compare our latent-space CoD with Stable-Diffusion-based codecs, including zero-shot DiffC~\cite{diffc}, DDCM~\cite{ddcm}, and fine-tuned PerCo (SD)~\cite{perco_sd} and OSCAR~\cite{oscar}. As shown in Figure~\ref{fig:Eval_Fondation_Model}, CoD achieves superior reconstruction quality. A more detailed quantitative comparison is presented in Section~\ref{sec:downstream}.

\subsection{Scaling Law}
\label{sec:eval_scaling}

Most diffusion codecs are constrained by the parameter count of Stable Diffusion (approximately 860M), making it difficult to study parameter scaling. It also remains unclear whether the performance advantages of diffusion codecs over GAN-based codecs (e.g., MS-ILLM~\cite{illm}) stem from architectural improvements or simply from model scaling. To explore this, we train pixel-space CoDs with varying channel dimensions at $256\times256$ resolution for 400k steps and evaluate them on Kodak. As shown in Figure~\ref{fig:Scaling_Law}, we observe a clear trend: increasing model parameters consistently enhances compression performance. Compared with the GAN-based MS-ILLM (181M parameters), CoD achieves notably better reconstruction quality even with only 49M parameters, confirming that the performance gain primarily originates from algorithmic improvements rather than model size.

\begin{figure*}[t]
  \centering
    \includegraphics[width=\linewidth]{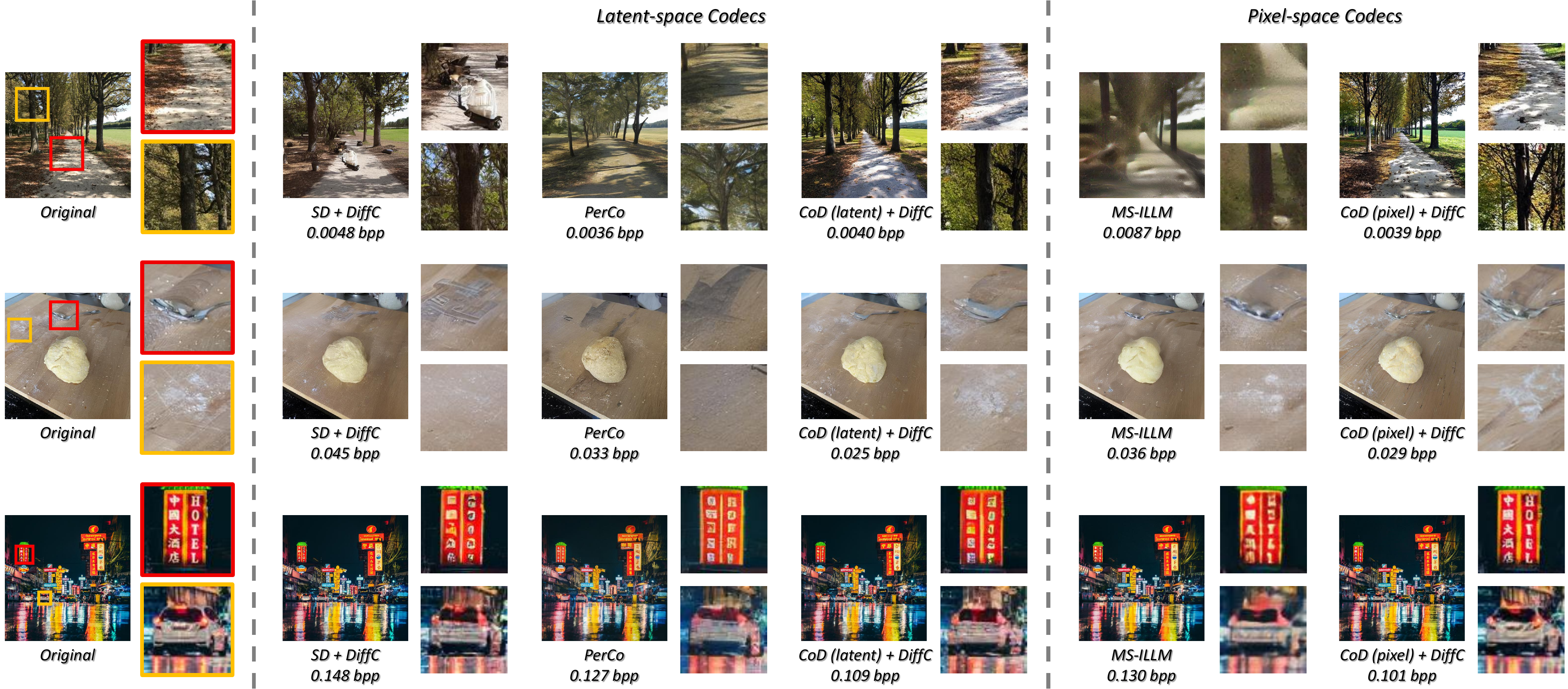}
    \vspace{-6mm}
    \caption{Visualization comparison on the CLIC2020 test set across bitrates from low (top) to high (bottom).} 
    \vspace{-3mm}
  \label{fig:Visual}
\end{figure*}

\subsection{Zero-Shot Distortion-Perception Control}

The proposed unified training jointly optimizes one-step distortion and multi-step perception, endowing CoD with the ability to control the distortion–perception trade-off directly through the number of sampling steps. As shown in Figure~\ref{fig:Compare_CoD_SD} (left), at 0.0039 bpp, pixel-space CoD attains the best perceptual quality at 25 steps and achieves a 3.4 dB PSNR improvement (from 16.2 dB to 19.6 dB) when reduced to a single step. Intermediate step counts smoothly interpolate between these two states. In contrast, diffusion codecs without this formulation (e.g., PerCo (SD)) fail to exhibit such controllable improvement through step reduction.

\subsection{Revisiting Pixel-Space Diffusion}

Recent diffusion codecs typically operate in the latent space of a KL-VAE, achieving impressive results at low bitrates (e.g., $<$0.1 bpp). However, their performance is fundamentally constrained by the entropy and reconstruction capacity of the underlying VAE. As shown in Figure~\ref{fig:Compare_CoD_SD} (right), DiffC~\cite{diffc} built on Stable Diffusion cannot surpass the reconstruction limit of the SD-VAE (PSNR 25.77 dB, LPIPS 0.069), with a bitrate ceiling around 0.26 bpp. Similarly, latent-space CoD remains bounded by VAE quality even up to 0.60 bpp, indicating that latent diffusion codecs inherently operate within a limited quality and bitrate range.

In contrast, pixel-space CoD directly models raw pixels to overcome this constraint. It can scale the bitrate beyond 1.0 bpp while continuously improving reconstruction quality. Pixel-space CoD surpasses latent diffusion models in LPIPS at around 0.3 bpp and maintains substantially higher PSNR across the entire range. This demonstrates the potential of pixel-space diffusion as a more general and unconstrained approach to learned compression in future research.

\begin{figure*}[t]
  \centering
    \includegraphics[width=\linewidth]{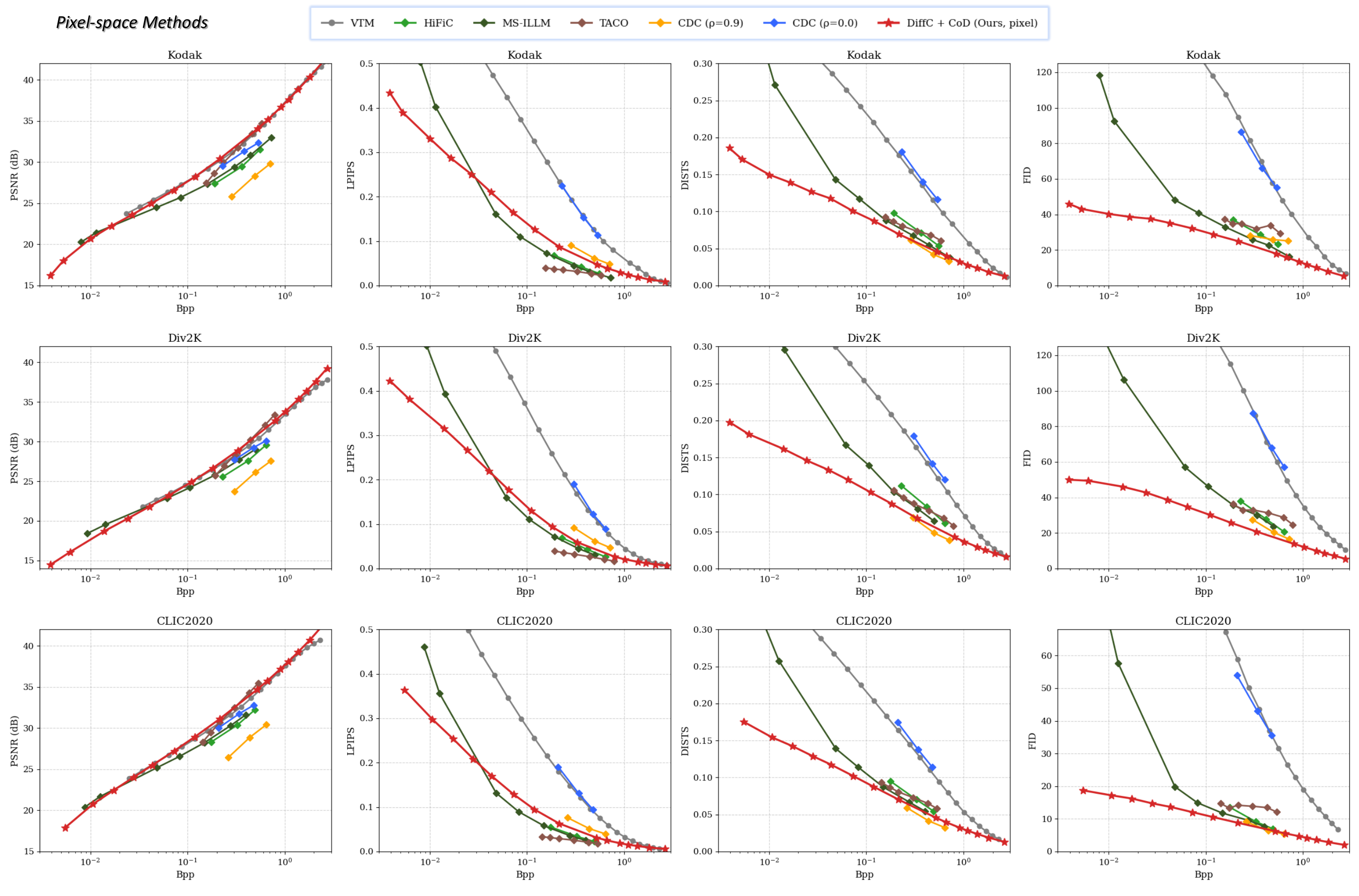}
    \vspace{-7mm}
    \caption{Comparison with pixel-space codecs. \textit{Note: HiFiC, MS-ILLM, CDC, and TACO are optimized using LPIPS, whereas CoD is not. So pixel-space CoD may not achieve the best LPIPS at certain bitrates, despite outperforming in PSNR, DISTS and FID.}} 
    \vspace{-3mm}
  \label{fig:RD_Pixel}
\end{figure*}

\begin{figure*}[t]
  \centering
    \includegraphics[width=\linewidth]{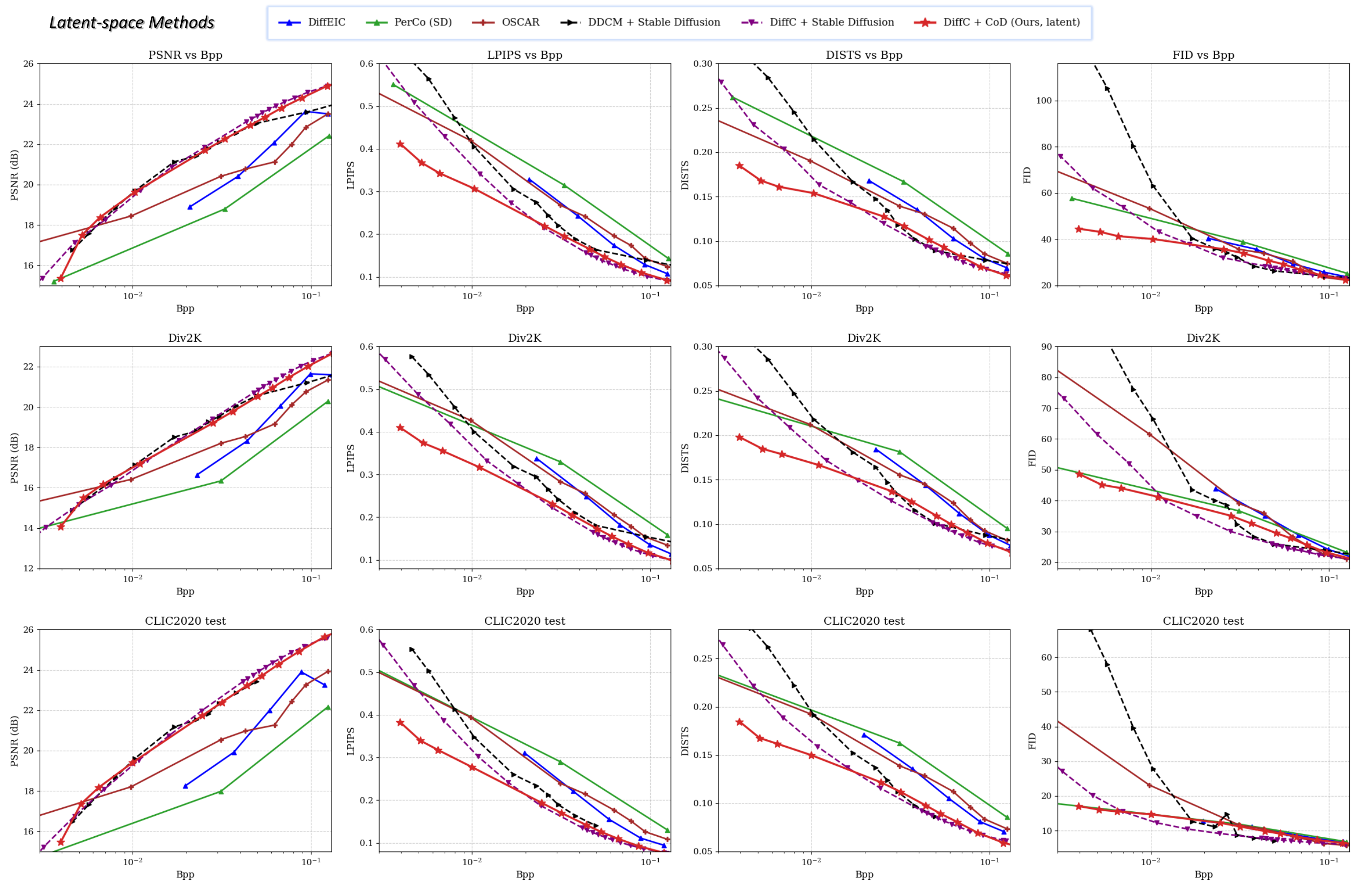}
    \vspace{-7mm}
    \caption{Comparison with latent-space codecs. Latent-space CoD demonstrates state-of-the-art performance towards low bitrates.} 
    \vspace{-3mm}
  \label{fig:RD_Latent}
\end{figure*}
\section{Downstream Compression Performance}
\label{sec:downstream}

In this section, we conduct a comprehensive evaluation of CoD when integrated into downstream compression frameworks. Since most diffusion-based codecs do not release their training code, we employ the zero-shot compression method DiffC~\cite{diffc} on CoD for comparison. 
% \textbf{\textcolor{red}{Further results on downstream codecs and applications are provided in the appendix, where finetuning one-step CoD (pixel) with DMD achieves real-time and SOTA performance (Section~\ref{sec:one-step}) and CoD serves effectively as a perceptual loss to significantly improving MS-ILLM~\cite{illm}.}}
\textcolor{darkred}{Additional downstream results are presented in the appendix, e.g., finetuning one-step CoD achieves real-time and SOTA performance (Section~\ref{sec:one-step}), and CoD-based perceptual loss that significantly enhances MS-ILLM~\cite{illm} performance (Section~\ref{sec:dmd_loss}).}

\noindent\textbf{Benchmarks and Baselines}.
Experiments are conducted on the Kodak~\cite{kodak}, CLIC2020 test set~\cite{clic} and Div2K~\cite{div2k}. All images are resized and center-cropped to a resolution of $512\times512$. We report distortion using PSNR and perceptual quality using LPIPS~\cite{lpips}, DISTS~\cite{dists}, and FID~\cite{fid}. Following~\cite{ddcm}, FID is computed on extracted patches ($64\times64$ for Kodak and $128\times128$ for CLIC2020). Bitrate is measured in bits per pixel (bpp). In addition to diffusion codecs (DiffC, DDCM, PerCo (SD), and OSCAR; see Section~\ref{sec:eval_method}), we further compare against the latent-space codec DiffEIC~\cite{diffeic} and pixel-space codecs including the traditional codec VTM~\cite{vvc} and perceptual codecs HiFiC~\cite{hific}, MS-ILLM~\cite{illm}, CDC~\cite{cdc} and TACO~\cite{taco}. We also discuss StableCodec~\cite{stablecodec} and OneDC~\cite{onedc} which are built on pretrained one-step diffusion models in the appendix.

\noindent\textbf{Pixel-space Comparison}. The rate-distortion curves on pixel space codecs are shown in Figure~\ref{fig:RD_Pixel}. Pixel space codecs are free of the limitation of VAE latents thus can achieve wider bitrates and quality ranges. Across all bitrates, our codec achieves superior PSNR, DISTS and FID values. Notably, perceptual codecs HiFiC, MS-ILLM, CDC ($\rho=0.9$) and TACO are optimized with LPIPS, whereas pixel-space CoD is not, which explains performance on LPIPS. 

Surprisingly, our codec achieves comparable BPP–PSNR performance to VTM (BD-Rate is $-2.1\%$ using VTM as anchor) while delivering significantly better perceptual quality. In contrast, HiFiC,  MS-ILLM and CDC ($\rho=0.9$) improve perceptual quality at the expense of PSNR. TACO removes the GAN~\cite{gan} loss to balance distortion and perception, but fails to achieve an optimal trade-off. By comparison, DiffC with pixel-space CoD attains the best balance across all bitrates. Although pixel diffusion has been less explored than latent diffusion, our results demonstrate that \textbf{CoD highlights the potential of pixel diffusion to achieve a superior rate–distortion–perception trade-off.}

\noindent\textbf{Latent-space Comparison}. The rate-distortion curves on latent space codecs are shown in Figure~\ref{fig:RD_Latent}. Compared to existing latent-space codecs, DiffC with latent-space CoD achieves the best reconstruction quality at lower bitrates (e.g., $<$0.02 bpp). At higher bitrates, the gap narrows as performance approaches the SD-VAE limit. At some points (e.g., FID at 0.02 bpp on CLIC), CoD performs slightly worse than DiffC due to smaller training scale, which can likely be closed with larger data and training in the future.

\noindent\textbf{Visualization}. Figure~\ref{fig:Visual} compares visual results across different bitrates among SD-based DiffC, PerCo (SD), and the pixel-space MS-ILLM. At low bitrates, CoD achieves higher fidelity than SD-based codecs and more realistic details than prior pixel-space methods. At high bitrates, latent codecs are constrained by the reconstruction quality of SD-VAE, e.g., small characters appear heavily distorted, while MS-ILLM also struggles to recover clear text. In contrast, pixel-space CoD reconstructs sharp and accurate details, demonstrating the strong potential of pixel-space diffusion.

\section{Related Works}
\label{sec:related}

\subsection{Generative Image Compression}

Unlike neural image codecs~\cite{balle2017end, hyperprior, elic} that primarily minimize reconstruction distortion (MSE), generative image compression~\cite{agustsson2019generative, iwai2021fidelity, po_elic, glc, multi_realism, dlf} aims to achieve both low distortion and high perceptual quality. Early works~\cite{hific, illm} introduced conditional GANs to enhance perceptual realism, successfully improving perceptual quality compared to MSE-optimized codecs, albeit at the cost of higher reconstruction distortion. TACO~\cite{taco} further incorporates textual information to mitigate MSE degradation, yet still falls short in achieving an ideal distortion–perception balance. To the best of our knowledge, we are the first to demonstrate that pixel-space diffusion can attain VTM-level distortion while delivering superior perceptual quality to GAN-based codecs.

\subsection{Diffusion Image Compression}

Recent advances in generative image compression increasingly leverage the powerful generative priors of diffusion models to achieve higher realism. PerCo~\cite{perco} compresses image representations and text captions as conditions, fine-tuning a pretrained latent diffusion model for perfect realism. In contrast, DiffEIC~\cite{diffeic} and OSCAR~\cite{oscar} show that removing the text bitstream can yield better overall performance. Other approaches~\cite{psc, idempotence} investigate zero-shot compression using pretrained diffusion models without fine-tuning. 
% For instance, posterior sampling~\cite{psc, idempotence} can be used to perform compression directly from the diffusion process. 
DiffC~\cite{diffc_original, diffc} introduces reverse-channel coding to progressively compress increasingly noised representations, while DDCM~\cite{ddcm} adopts random codebooks to encode the random sampled noises. Most of these methods depend on pretrained diffusion foundation models such as Stable Diffusion, which are theoretically suboptimal for compression. CoD offers an alternative, compression-oriented diffusion foundation model that can serve as a stronger backbone for such approaches, enabling superior performance.

Another early exploration on pixel-space diffusion codec is CDC~\cite{cdc}, which also formulates compression as conditions for diffusion. However, CoD differs in several key aspects. First, CoD emphasizes training at ultra-low bitrates ($\sim$0.0039 bpp) to strengthen its strong generative prior, and it relies purely on diffusion loss rather than traditional perceptual losses. In contrast, CDC is trained at high bitrates ($>$0.2 bpp), where dense information limits generative learning. Consequently, CDC requires perceptual losses such as LPIPS, otherwise its perceptual quality drops to the level of an MSE-optimized codec. Second, CoD embodies the concept of a foundation model, scaling training to push the frontier of compression research, while CDC does not consider scaling laws. Finally, CoD offers insights beyond the model itself: we provide a comprehensive analysis of scaling behavior, pixel- vs. latent-space and conditioning modalities, offering broader insights for future research.

\section{Conclusion and Limitation}
\label{sec:conclusion}

To conclude, we present CoD, the first compression-oriented diffusion foundation model. CoD is trained from scratch and enables end-to-end joint optimization of compression and diffusion generation, substantially improving diffusion-based compression performance, particularly at ultra-low bitrates. CoD is trained entirely on open datasets with low computational cost, facilitating reproducibility and further research. Building on CoD, we investigate the scaling laws of diffusion codecs and demonstrate the potential of pixel-space diffusion towards strong R–D–P trade-off across a wide bitrate range. We believe CoD establishes a solid foundation for future advances in diffusion-based compression.

Despite these advantages, CoD still has several limitations at this stage. First, it is not yet trained for high-resolution inputs. Although extending CoD to higher resolutions (e.g., 2K) is technically feasible, it would require substantially higher computational cost. As an early academic prototype, we leave this for future work. Second, similar to all diffusion codecs, CoD does not meet real-time coding requirements. Nevertheless, as shown in Section~\ref{sec:eval_scaling}, even a compact CoD model already surpasses GAN-based codecs by a large margin. With continued advances in diffusion distillation, we hope real-time will become attainable in the near future.

\section*{Change Log}
\label{sec:conclusion}

\begin{itemize}
    \setlength\itemsep{0.5em}
    \item \textbf{v1} \texttt{(2025-11-24)}: Initial submission.
    \item \textbf{v2} \texttt{(2025-12-10)}: Updated algorithms and results for finetuned one-step diffusion codec (App.~B), CoD-based perceptual supervision (App.~C), analysis on different downstream codecs (App.~D), and $\mathcal{X}$-prediction (App.~E).
    \item \textbf{v3} \texttt{(2026-03-16)}: Updated to CVPR camera ready version. Updated Github code link.
\end{itemize}
{
    \small
    \bibliographystyle{ieeenat_fullname}
    \bibliography{main}

@String(ICCV= {Int. Conf. Comput. Vis.})

@String(ICPR = {Int. Conf. Pattern Recog.})

@String(ICLR = {Int. Conf. Learn. Represent.})

@String(ICCV  = {ICCV})

@String(ICPR  = {ICPR})

@String(ICLR  = {ICLR})

@article{hific,
  title={High-fidelity generative image compression},
  author={Mentzer, Fabian and Toderici, George D and Tschannen, Michael and Agustsson, Eirikur},
  journal={Advances in neural information processing systems},
  volume={33},
  pages={11913--11924},
  year={2020}
}

@inproceedings{illm,
  title={Improving statistical fidelity for neural image compression with implicit local likelihood models},
  author={Muckley, Matthew J and El-Nouby, Alaaeldin and Ullrich, Karen and J{\'e}gou, Herv{\'e} and Verbeek, Jakob},
  booktitle={International Conference on Machine Learning},
  pages={25426--25443},
  year={2023},
  organization={PMLR}
}

@inproceedings{text+sketch,
title={Text + Sketch: Image Compression at Ultra Low Rates},
author={Eric Lei and Yigit Berkay Uslu and Hamed Hassani and Shirin Saeedi Bidokhti},
booktitle={ICML 2023 Workshop Neural Compression: From Information Theory to Applications},
year={2023},
}

@inproceedings{perco,
title={Towards image compression with perfect realism at ultra-low bitrates},
author={Marlene Careil and Matthew J. Muckley and Jakob Verbeek and St{\'e}phane Lathuili{\`e}re},
booktitle={The Twelfth International Conference on Learning Representations},
year={2024},
}

@article{perco_sd,
  author={Nikolai Körber and Eduard Kromer and Andreas Siebert and Sascha Hauke and Daniel Mueller-Gritschneder and Björn W. Schuller},
  title={PerCo (SD): Open Perceptual Compression},
  year={2024},
  journal={CoRR},
}

@article{oscar,
  title={OSCAR: One-Step Diffusion Codec Across Multiple Bit-rates},
  author={Guo, Jinpei and Ji, Yifei and Chen, Zheng and Liu, Kai and Liu, Min and Rao, Wang and Li, Wenbo and Guo, Yong and Zhang, Yulun},
  journal={arXiv preprint arXiv:2505.16091},
  year={2025}
}

@article{diffeic,
  title={Towards extreme image compression with latent feature guidance and diffusion prior},
  author={Li, Zhiyuan and Zhou, Yanhui and Wei, Hao and Ge, Chenyang and Jiang, Jingwen},
  journal={IEEE Transactions on Circuits and Systems for Video Technology},
  year={2024},
  publisher={IEEE}
}

@inproceedings{taco,
title={Neural Image Compression with Text-guided Encoding for both Pixel-level and Perceptual Fidelity},
author={Lee, Hagyeong and Kim, Minkyu and Kim, Jun-Hyuk and Kim, Seungeon and Oh, Dokwan and Lee, Jaeho},
booktitle={International Conference on Machine Learning},
year={2024}
}

@article{cdc,
  title={Lossy image compression with conditional diffusion models},
  author={Yang, Ruihan and Mandt, Stephan},
  journal={Advances in Neural Information Processing Systems},
  volume={36},
  pages={64971--64995},
  year={2023}
}

@article{diffc_original,
  title={Lossy compression with gaussian diffusion},
  author={Theis, Lucas and Salimans, Tim and Hoffman, Matthew D and Mentzer, Fabian},
  journal={arXiv preprint arXiv:2206.08889},
  year={2022}
}

@inproceedings{diffc,
title={Lossy Compression with Pretrained Diffusion Models},
author={Jeremy Vonderfecht and Feng Liu},
booktitle={The Thirteenth International Conference on Learning Representations},
year={2025},
}

@inproceedings{ddcm,
title={Compressed Image Generation with Denoising Diffusion Codebook Models},
author={Guy Ohayon and Hila Manor and Tomer Michaeli and Michael Elad},
booktitle={Forty-second International Conference on Machine Learning},
year={2025},
}

@InProceedings{stablecodec,
    author    = {Zhang, Tianyu and Luo, Xin and Li, Li and Liu, Dong},
    title     = {StableCodec: Taming One-Step Diffusion for Extreme Image Compression},
    booktitle = {Proceedings of the IEEE/CVF International Conference on Computer Vision (ICCV)},
    month     = {October},
    year      = {2025},
    pages     = {17379-17389}
}

@inproceedings{onedc,
title={One-Step Diffusion-Based Image Compression with Semantic Distillation},
author={Naifu Xue and Zhaoyang Jia and Jiahao Li and Bin Li and Yuan Zhang and Yan Lu},
booktitle={The Thirty-ninth Annual Conference on Neural Information Processing Systems},
year={2025},
}

@inproceedings{
    hyperprior,
    title={Variational image compression with a scale hyperprior},
    author={Johannes Ballé and David Minnen and Saurabh Singh and Sung Jin Hwang and Nick Johnston},
    booktitle={International Conference on Learning Representations},
    year={2018},
}

@misc{VVC,
    author = {{VVC-21.2}},
    year = {2023},
    title = {},
    howpublished = {\url{https://vcgit.hhi.fraunhofer.de/jvet/VVCSoftware_VTM/-/tree/VTM-21.2}. Accessed: 10/23/2023}    
}

@article{diffusion,
  title={Denoising diffusion probabilistic models},
  author={Ho, Jonathan and Jain, Ajay and Abbeel, Pieter},
  journal={Advances in neural information processing systems},
  volume={33},
  pages={6840--6851},
  year={2020}
}

@inproceedings{rf,
title={Flow Straight and Fast: Learning to Generate and Transfer Data with Rectified Flow},
author={Xingchao Liu and Chengyue Gong and qiang liu},
booktitle={The Eleventh International Conference on Learning Representations },
year={2023},
}

@article{gentiloni2024theoretical,
  title={Theoretical guarantees in kl for diffusion flow matching},
  author={Gentiloni Silveri, Marta and Durmus, Alain and Conforti, Giovanni},
  journal={Advances in Neural Information Processing Systems},
  volume={37},
  pages={138432--138473},
  year={2024}
}

@article{v-pred,
  title={Progressive distillation for fast sampling of diffusion models},
  author={Salimans, Tim and Ho, Jonathan},
  journal={arXiv preprint arXiv:2202.00512},
  year={2022}
}

@article{jit,
  title={Back to Basics: Let Denoising Generative Models Denoise},
  author={Li, Tianhong and He, Kaiming},
  journal={arXiv preprint arXiv:2511.13720},
  year={2025}
}

@inproceedings{SD,
  title={High-resolution image synthesis with latent diffusion models},
  author={Rombach, Robin and Blattmann, Andreas and Lorenz, Dominik and Esser, Patrick and Ommer, Bj{\"o}rn},
  booktitle={Proceedings of the IEEE/CVF conference on computer vision and pattern recognition},
  pages={10684--10695},
  year={2022}
}

@inproceedings{pixart_alpha,
title={PixArt-\${\textbackslash}alpha\$: Fast Training of Diffusion Transformer for Photorealistic Text-to-Image Synthesis},
author={Junsong Chen and Jincheng YU and Chongjian GE and Lewei Yao and Enze Xie and Zhongdao Wang and James Kwok and Ping Luo and Huchuan Lu and Zhenguo Li},
booktitle={The Twelfth International Conference on Learning Representations},
year={2024},
}

@article{ddt,
  title={DDT: Decoupled diffusion transformer},
  author={Wang, Shuai and Tian, Zhi and Huang, Weilin and Wang, Limin},
  journal={arXiv preprint arXiv:2504.05741},
  year={2025}
}

@article{pixnerd,
  title={Pixnerd: Pixel neural field diffusion},
  author={Wang, Shuai and Gao, Ziteng and Zhu, Chenhui and Huang, Weilin and Wang, Limin},
  journal={arXiv preprint arXiv:2507.23268},
  year={2025}
}

@inproceedings{repa,
title={Representation Alignment for Generation: Training Diffusion Transformers Is Easier Than You Think},
author={Sihyun Yu and Sangkyung Kwak and Huiwon Jang and Jongheon Jeong and Jonathan Huang and Jinwoo Shin and Saining Xie},
booktitle={The Thirteenth International Conference on Learning Representations},
year={2025},
}

@article{score_matching,
  title={Generative modeling by estimating gradients of the data distribution},
  author={Song, Yang and Ermon, Stefano},
  journal={Advances in neural information processing systems},
  volume={32},
  year={2019}
}

@inproceedings{cfg,
title={Classifier-Free Diffusion Guidance},
author={Jonathan Ho and Tim Salimans},
booktitle={NeurIPS 2021 Workshop on Deep Generative Models and Downstream Applications},
year={2021},
}

@inproceedings{dmd,
  title={One-step diffusion with distribution matching distillation},
  author={Yin, Tianwei and Gharbi, Micha{\"e}l and Zhang, Richard and Shechtman, Eli and Durand, Fredo and Freeman, William T and Park, Taesung},
  booktitle={Proceedings of the IEEE/CVF conference on computer vision and pattern recognition},
  pages={6613--6623},
  year={2024}
}

@inproceedings{blip,
  title={Blip-2: Bootstrapping language-image pre-training with frozen image encoders and large language models},
  author={Li, Junnan and Li, Dongxu and Savarese, Silvio and Hoi, Steven},
  booktitle={International conference on machine learning},
  pages={19730--19742},
  year={2023},
  organization={PMLR}
}

@inproceedings{vqgan,
  title={Taming transformers for high-resolution image synthesis},
  author={Esser, Patrick and Rombach, Robin and Ommer, Bjorn},
  booktitle={Proceedings of the IEEE/CVF conference on computer vision and pattern recognition},
  pages={12873--12883},
  year={2021}
}

@inproceedings{fsq,
title={Finite Scalar Quantization: {VQ}-{VAE} Made Simple},
author={Fabian Mentzer and David Minnen and Eirikur Agustsson and Michael Tschannen},
booktitle={The Twelfth International Conference on Learning Representations},
year={2024},
}

@article{imagenet,
  title={Imagenet large scale visual recognition challenge},
  author={Russakovsky, Olga and Deng, Jia and Su, Hao and Krause, Jonathan and Satheesh, Sanjeev and Ma, Sean and Huang, Zhiheng and Karpathy, Andrej and Khosla, Aditya and Bernstein, Michael and others},
  journal={International journal of computer vision},
  volume={115},
  number={3},
  pages={211--252},
  year={2015},
  publisher={Springer}
}

@article{openimage,
  title={The open images dataset v4: Unified image classification, object detection, and visual relationship detection at scale},
  author={Kuznetsova, Alina and Rom, Hassan and Alldrin, Neil and Uijlings, Jasper and Krasin, Ivan and Pont-Tuset, Jordi and Kamali, Shahab and Popov, Stefan and Malloci, Matteo and Kolesnikov, Alexander and others},
  journal={International journal of computer vision},
  volume={128},
  number={7},
  pages={1956--1981},
  year={2020},
  publisher={Springer}
}

@inproceedings{sa1b,
  title={Segment anything},
  author={Kirillov, Alexander and Mintun, Eric and Ravi, Nikhila and Mao, Hanzi and Rolland, Chloe and Gustafson, Laura and Xiao, Tete and Whitehead, Spencer and Berg, Alexander C and Lo, Wan-Yen and others},
  booktitle={Proceedings of the IEEE/CVF international conference on computer vision},
  pages={4015--4026},
  year={2023}
}

@misc{kodak,
  title        = {Kodak Lossless True Color Image Suite},
  author       = {{Eastman Kodak Company}},
  year         = {1999},
  howpublished = {\url{http://r0k.us/graphics/kodak/}},
  note         = {Accessed: 2025-11-08}
}

@article{clic,
  title={Clic 2020: Challenge on learned image compression},
  author={Toderici, George and Theis, Lucas and Johnston, Nick and Agustsson, Eirikur and Mentzer, Fabian and Ball{\'e}, Johannes and Shi, Wenzhe and Timofte, Radu},
  journal={Retrieved March},
  volume={29},
  pages={2021},
  year={2020}
}

@inproceedings{div2k,
  title={Ntire 2017 challenge on single image super-resolution: Dataset and study},
  author={Agustsson, Eirikur and Timofte, Radu},
  booktitle={Proceedings of the IEEE conference on computer vision and pattern recognition workshops},
  pages={126--135},
  year={2017}
}

@inproceedings{resnet,
  title={Deep residual learning for image recognition},
  author={He, Kaiming and Zhang, Xiangyu and Ren, Shaoqing and Sun, Jian},
  booktitle={Proceedings of the IEEE conference on computer vision and pattern recognition},
  pages={770--778},
  year={2016}
}

@article{transformer,
  title={Attention is all you need},
  author={Vaswani, Ashish and Shazeer, Noam and Parmar, Niki and Uszkoreit, Jakob and Jones, Llion and Gomez, Aidan N and Kaiser, {\L}ukasz and Polosukhin, Illia},
  journal={Advances in neural information processing systems},
  volume={30},
  year={2017}
}

@inproceedings{rdp,
  title={Rethinking lossy compression: The rate-distortion-perception tradeoff},
  author={Blau, Yochai and Michaeli, Tomer},
  booktitle={International Conference on Machine Learning},
  pages={675--685},
  year={2019},
  organization={PMLR}
}

@article{gan,
  title={Generative adversarial networks},
  author={Goodfellow, Ian and Pouget-Abadie, Jean and Mirza, Mehdi and Xu, Bing and Warde-Farley, David and Ozair, Sherjil and Courville, Aaron and Bengio, Yoshua},
  journal={Communications of the ACM},
  volume={63},
  number={11},
  pages={139--144},
  year={2020},
  publisher={ACM New York, NY, USA}
}

@inproceedings{lpips,
  title={The unreasonable effectiveness of deep features as a perceptual metric},
  author={Zhang, Richard and Isola, Phillip and Efros, Alexei A and Shechtman, Eli and Wang, Oliver},
  booktitle={Proceedings of the IEEE conference on computer vision and pattern recognition},
  pages={586--595},
  year={2018}
}

@article{dists,
  title={Image quality assessment: Unifying structure and texture similarity},
  author={Ding, Keyan and Ma, Kede and Wang, Shiqi and Simoncelli, Eero P},
  journal={IEEE transactions on pattern analysis and machine intelligence},
  volume={44},
  number={5},
  pages={2567--2581},
  year={2020},
  publisher={IEEE}
}

@article{fid,
  title={Gans trained by a two time-scale update rule converge to a local nash equilibrium},
  author={Heusel, Martin and Ramsauer, Hubert and Unterthiner, Thomas and Nessler, Bernhard and Hochreiter, Sepp},
  journal={Advances in neural information processing systems},
  volume={30},
  year={2017}
}

@article{dino_v2,
title={{DINO}v2: Learning Robust Visual Features without Supervision},
author={Oquab, Maxime and Darcet, Timoth{\'e}e and Moutakanni, Th{\'e}o and Vo, Huy and Szafraniec, Marc and Khalidov, Vasil and Fernandez, Pierre and Haziza, Daniel and Massa, Francisco and El-Nouby, Alaaeldin and others},
journal={Transactions on Machine Learning Research},
year={2024},
}

@article{lora,
  title={Lora: Low-rank adaptation of large language models.},
  author={Hu, Edward J and Shen, Yelong and Wallis, Phillip and Allen-Zhu, Zeyuan and Li, Yuanzhi and Wang, Shean and Wang, Lu and Chen, Weizhu and others},
  journal={ICLR},
  volume={1},
  number={2},
  pages={3},
  year={2022}
}

@article{max_score_norm,
  title={Optimizing Input of Denoising Score Matching is Biased Towards Higher Score Norm},
  author={Xu, Tongda},
  journal={arXiv preprint arXiv:2511.11727},
  year={2025}
}

@inproceedings{agustsson2019generative,
  title={Generative adversarial networks for extreme learned image compression},
  author={Agustsson, Eirikur and Tschannen, Michael and Mentzer, Fabian and Timofte, Radu and Gool, Luc Van},
  booktitle={Proceedings of the IEEE/CVF International Conference on Computer Vision},
  pages={221--231},
  year={2019}
}

@inproceedings{iwai2021fidelity,
  title={Fidelity-controllable extreme image compression with generative adversarial networks},
  author={Iwai, Shoma and Miyazaki, Tomo and Sugaya, Yoshihiro and Omachi, Shinichiro},
  booktitle={2020 25th International Conference on Pattern Recognition (ICPR)},
  pages={8235--8242},
  year={2021},
  organization={IEEE}
}

@inproceedings{po_elic,
  title={PO-ELIC: Perception-Oriented Efficient Learned Image Coding},
  author={He, Dailan and Yang, Ziming and Yu, Hongjiu and Xu, Tongda and Luo, Jixiang and Chen, Yuan and Gao, Chenjian and Shi, Xinjie and Qin, Hongwei and Wang, Yan},
  booktitle={Proceedings of the IEEE/CVF Conference on Computer Vision and Pattern Recognition},
  pages={1764--1769},
  year={2022}
}

@inproceedings{multi_realism,
  title={Multi-realism image compression with a conditional generator},
  author={Agustsson, Eirikur and Minnen, David and Toderici, George and Mentzer, Fabian},
  booktitle={Proceedings of the IEEE/CVF Conference on Computer Vision and Pattern Recognition},
  pages={22324--22333},
  year={2023}
}

@inproceedings{glc,
  title={Generative latent coding for ultra-low bitrate image compression},
  author={Jia, Zhaoyang and Li, Jiahao and Li, Bin and Li, Houqiang and Lu, Yan},
  booktitle={Proceedings of the IEEE/CVF Conference on Computer Vision and Pattern Recognition},
  pages={26088--26098},
  year={2024}
}

@InProceedings{dlf,
  author={Xue, Naifu and Jia, Zhaoyang and Li, Jiahao and Li, Bin and Zhang, Yuan and Lu, Yan},
  title={DLF: Extreme Image Compression with Dual-generative Latent Fusion},
  booktitle={Proceedings of the IEEE/CVF International Conference on Computer Vision (ICCV)},
  month = {Oct},
  year={2025},
}

@inproceedings{psc,
title={{PSC}: Posterior Sampling-Based Compression},
author={Noam Elata and Tomer Michaeli and Michael Elad},
booktitle={15th International Conference on Sampling Theory and Applications},
year={2025},
}

@inproceedings{idempotence,
title={Idempotence and Perceptual Image Compression},
author={Tongda Xu and Ziran Zhu and Dailan He and Yanghao Li and Lina Guo and Yuanyuan Wang and Zhe Wang and Hongwei Qin and Yan Wang and Jingjing Liu and Ya-Qin Zhang},
booktitle={The Twelfth International Conference on Learning Representations},
year={2024},
}

@inproceedings{balle2017end,
  title={End-to-end optimized image compression},
  author={Ball{\'e}, Johannes and Laparra, Valero and Simoncelli, Eero P},
  booktitle={5th International Conference on Learning Representations, ICLR 2017},
  year={2017}
}

@inproceedings{elic,
  title={Elic: Efficient learned image compression with unevenly grouped space-channel contextual adaptive coding},
  author={He, Dailan and Yang, Ziming and Peng, Weikun and Ma, Rui and Qin, Hongwei and Wang, Yan},
  booktitle={Proceedings of the IEEE/CVF Conference on Computer Vision and Pattern Recognition},
  pages={5718--5727},
  year={2022}
}
}

\newpage
\appendix
\noindent{\huge \textbf{Appendices}}\\

This document provides the supplementary material to \textbf{Co}mpression-oriented \textbf{D}iffusion foundation model, \textbf{CoD}. Beyond additional training details and extended results, we further investigate its characteristics and explore additional downstream codec applications.

The key conclusions are summarized below:

\begin{itemize}
\item \textbf{Extreme 64-bit compression}: CoD compresses image into only 64 bits while preserving correct semantics.
\item \textbf{One-step diffusion codec}: Finetuning one-step CoD achieves real-time and SOTA performance, comparable to StableCodec~\cite{stablecodec} and OneDC~\cite{onedc}.
\item \textbf{CoD as perceptual loss}: CoD-based perceptual loss significantly improves MS-ILLM~\cite{illm}.
\item \textbf{$\mathcal{X}$-prediction advantage}: Using $\mathcal{X}$-prediction yields better performance than $\mathcal{V}$-prediction for CoD (pixel).
\end{itemize}

\section{Towards 64-Bit Image Compression}
\label{sec:64bits}

In this paper, we primarily discuss CoD at 0.0039 bpp, which corresponds to 1024 bits for a $512\times512$ image. Results demonstrate that the shape, color, and high-level semantic fidelity are well preserved under this bitrate. In this section, we further explore a more extreme compression of 64 bits for a $512\times512$ image (i.e., 0.00024 bpp) to analyze the boundary of semantic-level compression. To achieve 64 bits, our encoder downsamples the original image to $1/128$ of its original resolution ($4\times4$ patches) and uses a codebook size of $2^{4} = 16$. Since latent-space CoD performs better at lower bitrates (Figure~5 in the paper), we train our 64-bit CoD on latent space.

\begin{figure}[h]
  \centering
    \includegraphics[width=\linewidth]{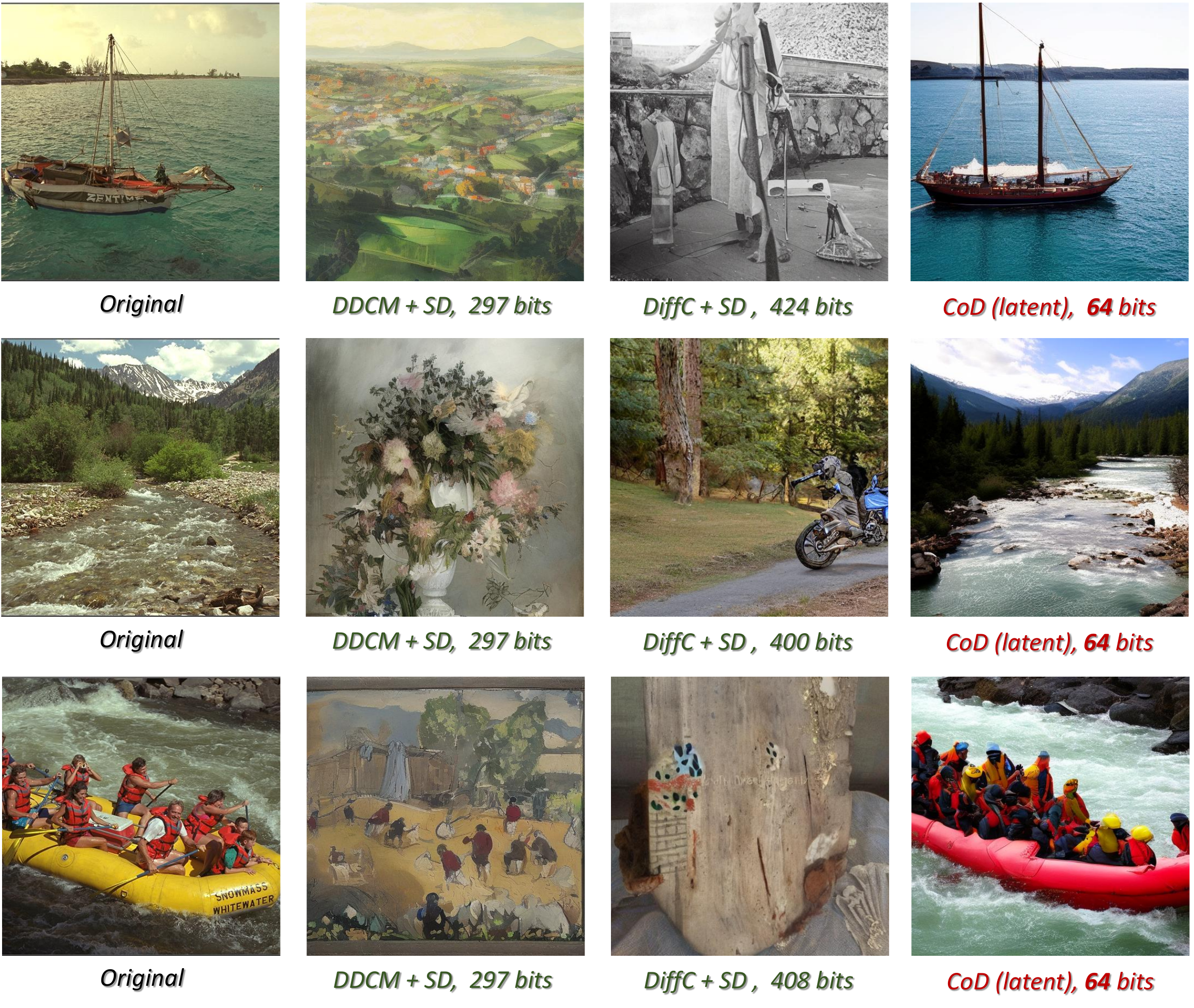}
    \vspace{-7mm}
    \caption{Evaluation of 64-bit CoD on Kodak at $512\times512$.} 
    \vspace{-3mm}
  \label{fig:64bits_Visual}
\end{figure}

\begin{figure*}[t]
  \centering
    \includegraphics[width=\linewidth]{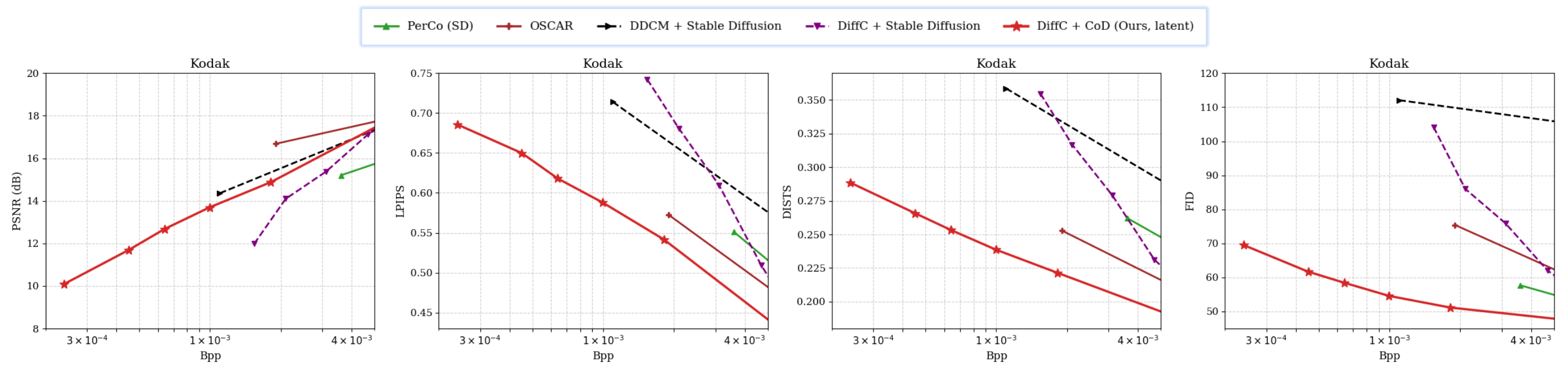}
    \vspace{-7mm}
    \caption{Rate-Distortion curves for 64-bit CoD and Stable Diffusion based codecs on Kodak at $512\times512$.} 
    \vspace{-3mm}
  \label{fig:64bit_RD}
\end{figure*}

\begin{figure*}[t]
  \centering
    \includegraphics[width=\linewidth]{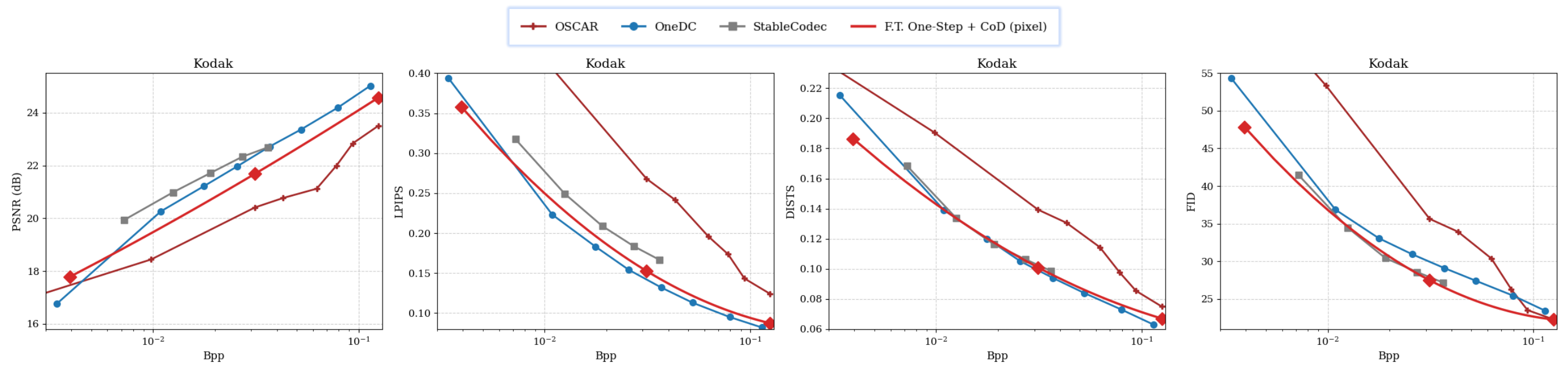}
    \vspace{-7mm}
    \caption{Rate-Distortion curves for finetuned one-step CoD and other one-step diffusion codecs on Kodak at $512\times512$.} 
    \vspace{-1mm}
  \label{fig:OneStep_RD}
\end{figure*}

We evaluate our 64-bit CoD on the Kodak dataset in Figure~\ref{fig:64bits_Visual}. Surprisingly, under such an extremely low bitrate, CoD successfully reconstructs the correct semantics of the original image. By contrast, Stable Diffusion based codecs DDCM~\cite{ddcm} and DiffC~\cite{diffc} fail to reconstruct correct semantics even at a fourfold ($4\times$) higher bit cost. We further leverage DiffC to evaluate the downstream compression performance on Kodak~\cite{kodak} in Figure~\ref{fig:64bit_RD}, where CoD-based DiffC significantly outperforms other Stable Diffusion based codecs. Our scheme achieves a FID of 70 with only less than $10\%$ of the bits of prior codecs, highlighting the extreme compression capability of 64-bit CoD.

\begin{figure}[h]
  \centering
    % \vspace{-1mm}
    \includegraphics[width=\linewidth]{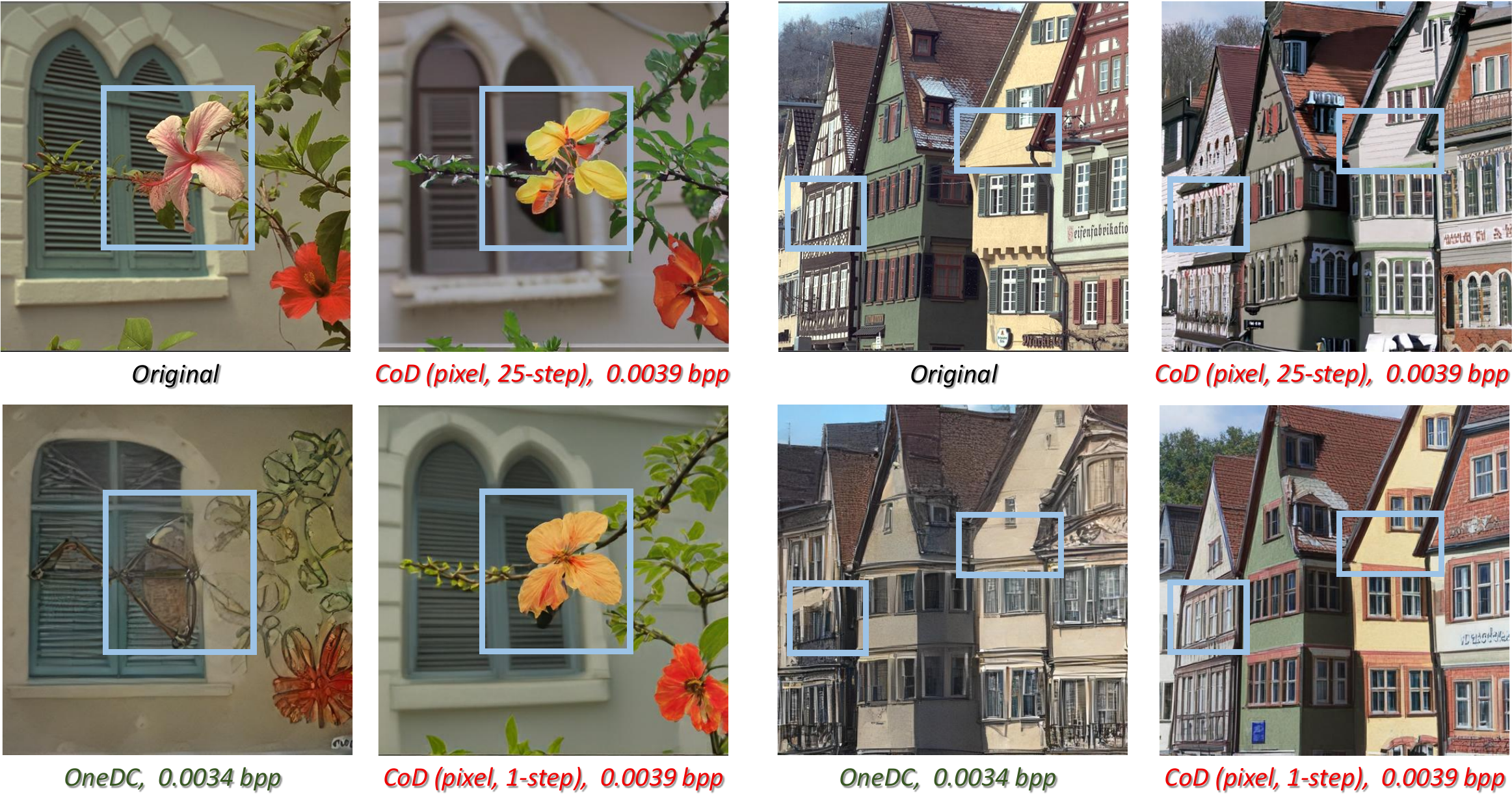}
    \vspace{-6mm}
    \caption{Evaluation of one-step CoD on Kodak at $512\times512$.} 
    \vspace{-4mm}
  \label{fig:OneStep_Visual}
\end{figure}

\begin{figure*}[t]
  \centering
    \includegraphics[width=\linewidth]{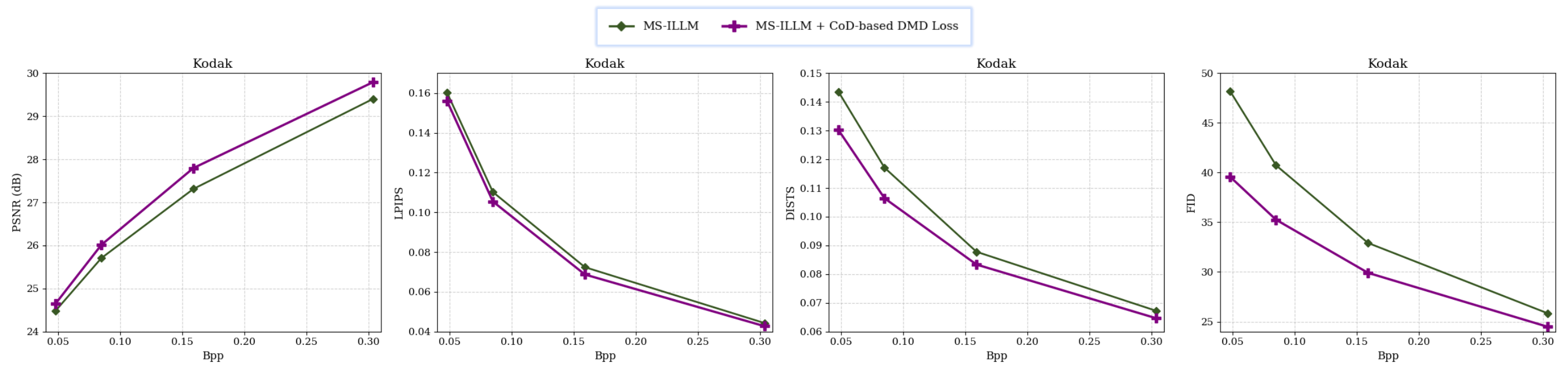}
    \vspace{-7mm}
    \caption{Rate-Distortion curves for MS-ILLM~\cite{illm} and finetuning MS-ILLM decoder with CoD-based DMD loss on Kodak at $512\times512$.} 
    \vspace{-3mm}
  \label{fig:DMD_loss_RD}
\end{figure*}

\section{Towards One-Step Diffusion Compression}
\label{sec:one-step}

\subsection{Preliminary}

Recently, one-step diffusion models have demonstrated effectiveness in generative image compression. Unlike multi-step diffusion, which optimizes each diffusion timestep separately, one-step diffusion can be trained end-to-end, enabling better overall performance. OSCAR~\cite{oscar} fine-tuned a multi-step Stable Diffusion to function as a one-step image codec. StableCodec~\cite{stablecodec} leverages a one-step SD-Turbo as the foundation model, achieving significantly improved performance. Similarly, OneDC~\cite{onedc} employs a one-step DMD-distilled~\cite{dmd} Stable Diffusion as the foundation model for one-step diffusion compression. The success of these approaches motivates our exploration of CoD in the context of one-step diffusion foundation models.

Since the SD-Turbo distillation process is not fully public, we follow DMD~\cite{dmd} to distill our diffusion model. Given a one-step diffusion network $G_\theta(z)$ that generates $x$ using random noise $z$, DMD proves that the KL divergence between the real and fake distributions can be expressed as a distribution-matching loss:
\begin{equation}
\nabla_\theta D_{KL} = \mathbb{E}_{\substack{z \sim \mathcal{N}(0; \mathbf{I}) \\ x = G_\theta(z)}} \left[ - \left( s_{\text{real}}(x) - s_{\text{fake}}(x) \right) \frac{dG}{d\theta} \right],
\label{equ:dmd}
\end{equation}
where $s_{\mathrm{real}}(x)$ and $s_{\mathrm{fake}}(x)$ are the score functions~\cite{score_matching} of their respective distributions. By adding random noise to $x$, the score can be predicted by a diffusion model. DMD leverages a pretrained multi-step diffusion model for the real score and dynamically trains a diffusion model to estimate the fake score for one-step samples $x$.
 
\subsection{Distilling a CoD}

We train a one-step CoD using a combination of losses and multi-step CoD-based DMD loss:
\begin{equation}
\mathcal{L_{\mathrm{OS}}} = \mathcal{L}_{\mathrm{pixel}}
+ \mathcal{L}_{\mathrm{perc}}
+ \lambda \,\cdot\, \mathcal{L}_{\mathrm{REPA}}
+ \beta \,\cdot\, \mathcal{L}_{\mathrm{C}},
\end{equation}
where $L_{\mathrm{pixel}}= \mathcal{L}_{1} 
+ \mathcal{L}_{\mathrm{LPIPS}}$ consists of pixel-space L1 loss and the LPIPS~\cite{lpips} loss, 
$L_{\mathrm{perc}}= 2 \,\cdot\, \mathcal{L}_{\mathrm{DMD}} 
+ 0.01 \,\cdot\, \mathcal{L}_{\mathrm{GAN}}$ consists of a DMD loss and a patch GAN loss~\cite{vqgan}, 
$\mathcal{L}_{\mathrm{REPA}}$ is the representation alignment loss and $\mathcal{L}_{\mathrm{C}}$ is th codebook commitment loss. We set $\lambda=0.5$ and $\beta=0.25$.

For simplicity, we optimize only the pixel-space CoD, allowing pixel-space losses to be computed directly. Following the DMD procedure, we update the one-step CoD every 10 steps, while the remaining 9 steps train the fake score diffusion model. The model is trained for 100K steps (i.e., 10K steps for the one-step CoD) with a batch size of 32, which takes approximately 96 A100 GPU hours in total, which is much less than 1664 A100 GPU hours of DMD-based Stable Diffusion distillation. The learning rates for one-step CoD and fake score network are set to $10^{-5}$. 

\subsection{Finetuning to Higher Bitrates}

In the previous section, we distilled the one-step CoD model at 0.0039 bpp. We further observe that the distilled model can be efficiently adapted to higher bitrates. Following PerCo and OSCAR, we adjust both the downsampling scale in the condition encoder and the codebook size according to the target bitrate. Specifically, we train two variants: 0.0312 bpp with $16\times$ downsampling and codebook size 256, and 0.125 bpp with $8\times$ downsampling and codebook size 256.

\noindent\textbf{Stage I: initialization}. 
We randomly set the parameters of the condition encoder, codebook, and condition decoder, while loading the pre-trained multi-step CoD weights for the diffusion module. During this warm-up phase, the diffusion module is trained using LoRA~\cite{lora} with rank 32. To accelerate convergence, we remove the $L_{\mathrm{perc}}$ term from $L_{\mathrm{OS}}$ and train for 200K steps with a learning rate of $10^{-4}$.

\noindent\textbf{Stage II: Fine-tuning}. 
We then fine-tune all model components jointly using the full $L_{\mathrm{OS}}$ for another 100K steps with a batch size of 32 and a learning rate of $10^{-5}$. With perceptual supervision restored, the fine-tuned one-step CoD achieves significantly improved performance at higher bitrates.

\subsection{Evaluation}

\noindent\textbf{Comparison with one-step codecs}.
We evaluate the fine-tuned one-step CoD on the Kodak dataset (Figure~\ref{fig:OneStep_RD}). It yields competitive reconstruction quality compared with SOTA one-step diffusion codecs such as OneDC and StableCodec, particularly at ultra-low bitrates. Notably, our method does not rely on any pretrained components from Stable Diffusion (e.g., one-step models or SD-VAEs) to reach this performance. We currently adopt fixed-length coding for the codebook indices for simplicity. Further gains can be achieved by introducing an entropy model or applying latent compression~\cite{glc}. Visual comparisons in Figure~\ref{fig:OneStep_Visual} further show that one-step CoD preserves higher fidelity than both multi-step CoD and OneDC.

\noindent\textbf{Complexity Analysis}. 
DiT injects conditios and timesteps through AdaLN-Zero layers, which account for more than 200M parameters in CoD. However, CoD concatenates compression conditions directly with the noised input along the channel dimension instead of injecting them through AdaLN-Zero (see Section~\ref{sec:pipeline}). Under the one-step setting, the timestep is fixed to 0, making all AdaLN-Zero outputs constant and therefore redundant. Thus, we precompute these constants and remove the AdaLN-Zero modules to further reduce computation and memory access. As shown in Table~\ref{tab:complexity}, one-step CoD processes a $512\times512$ image in only 25.2 ms, which is even faster than a single step of Stable Diffusion used in OneDC and StableCodec. This demonstrates its potential in real-time application.

\section{Towards Perceptual Supervision via CoD}
\label{sec:dmd_loss}

In the DMD loss in Equation~\ref{equ:dmd}, for any input $x$, we can optimize its KL divergence using CoD to estimate real and fake scores. This indicates that CoD can act as a perceptual supervision mechanism, enhancing the realism of reconstructions for a variety of networks, including other pixel-space codecs, restoration models, or super-resolution models. To demonstrate this potential, we finetune the pixel-space codec MS-ILLM~\cite{illm} using the CoD-based DMD loss.

\begin{table}[h]
\centering
\resizebox{\linewidth}{!}{
    \begin{tabular}{l|cccc}
    \toprule
    Methods & PSNR & LPIPS & DISTS & FID \\
    \midrule
    MS-ILLM & \textbf{21.43 dB} & 0.403 & 0.271 & 92.5\\
    % \midrule
    + CoD-based DMD loss & 21.08 dB & \textbf{0.376} & \textbf{0.248} & \textbf{80.92}\\
    \bottomrule
    \end{tabular}
    \vspace{1mm}
}
\vspace{-2mm}
\caption{Finetuning the decoder of MS-ILLM using CoD as a perceptual supervision. Evaluated on Kodak at $512\times512$. BPP=0.011.}
\label{tab:cod_dmd_loss}
\end{table}

\begin{figure}[h]
  \centering
    \includegraphics[width=\linewidth]{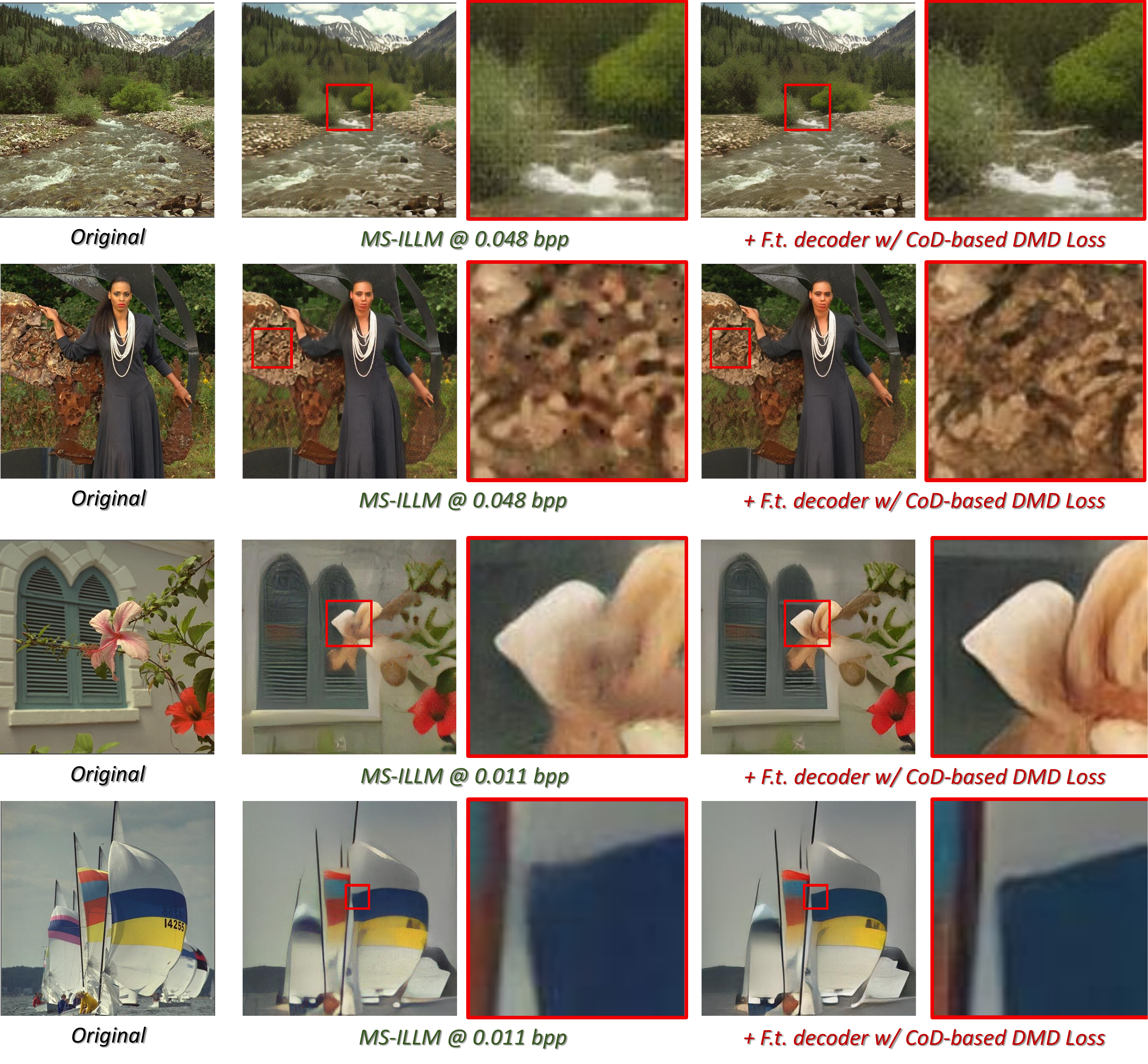}
    \vspace{-7mm}
    \caption{Finetuning MS-ILLM decoder with CoD-based DMD loss.} 
    \vspace{-3mm}
  \label{fig:DMD_loss_Visual}
\end{figure}

\begin{figure*}[t]
  \centering
    \includegraphics[width=\linewidth]{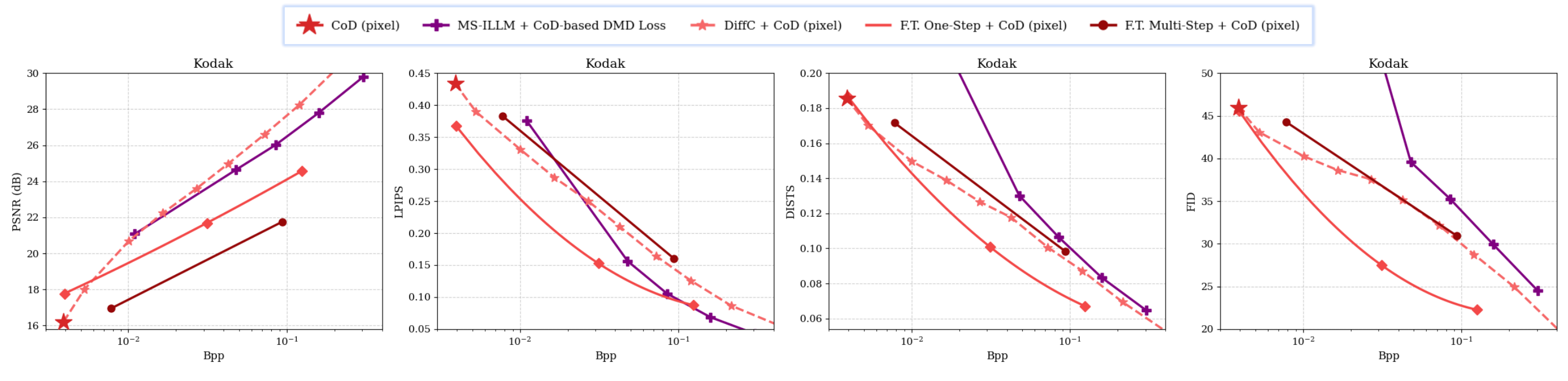}
    \vspace{-7mm}
    \caption{Rate-Distortion curves for different downstream codecs built on pixel-space CoD on Kodak at $512\times512$.} 
    \vspace{-3mm}
  \label{fig:Compare_downstream_RD}
\end{figure*}

\noindent\textbf{Evaluation}.
MS-ILLM is typically pretrained using the MSE loss, after which the encoder and entropy model are fixed while the decoder is finetuned with perceptual losses such as LPIPS and GAN~\cite{gan} loss. Following this paradigm, we finetune only the decoder of MS-ILLM, replacing the GAN loss with the CoD-based DMD loss. Remarkably, after optimizing the decoder for just 2K steps with a batch size of 32 (20K steps total including fake score learning), MS-ILLM demonstrates substantially improved perceptual quality across all metrics, as shown in Figure~\ref{fig:DMD_loss_RD}. Towards ultra-low bitrates like 0.011 bpp where the information is severely distorted, we find increasing the learning rate of fake score network training can better capture the distorted distribution to enhance realism, as shown in Table~\ref{tab:cod_dmd_loss}. Several visual examples are presented in Figure~\ref{fig:DMD_loss_Visual}, where the CoD-based DMD loss markedly reduces artifacts and yields clearer edges and finer details. Since the encoder is fixed, the encoded information remains heavily distorted under MSE-only optimization, resulting in poor overall reconstruction. We expect that jointly tuning the entire network with the CoD-based DMD loss could further improve perceptual quality.

\noindent\textbf{Discussion}.
Instead of relying on text conditions, CoD learns native image conditions for diffusion. Under DMD supervision, conditions are important since they guide which types of realistic details to generate. Native image conditions provide directions that more closely align with the original image, making CoD theoretically better suited as DMD supervision for reconstruction-related tasks. Moreover, CoD is text-free, eliminating the need for extensive captioning at each optimization step and thereby improving training efficiency. In addition, adopting latent diffusion for DMD loss is non-trivial for pixel-space codecs, whereas our pixel-space CoD offers a more direct and compatible solution.

\begin{figure*}[t]
  \centering
    \includegraphics[width=\linewidth]{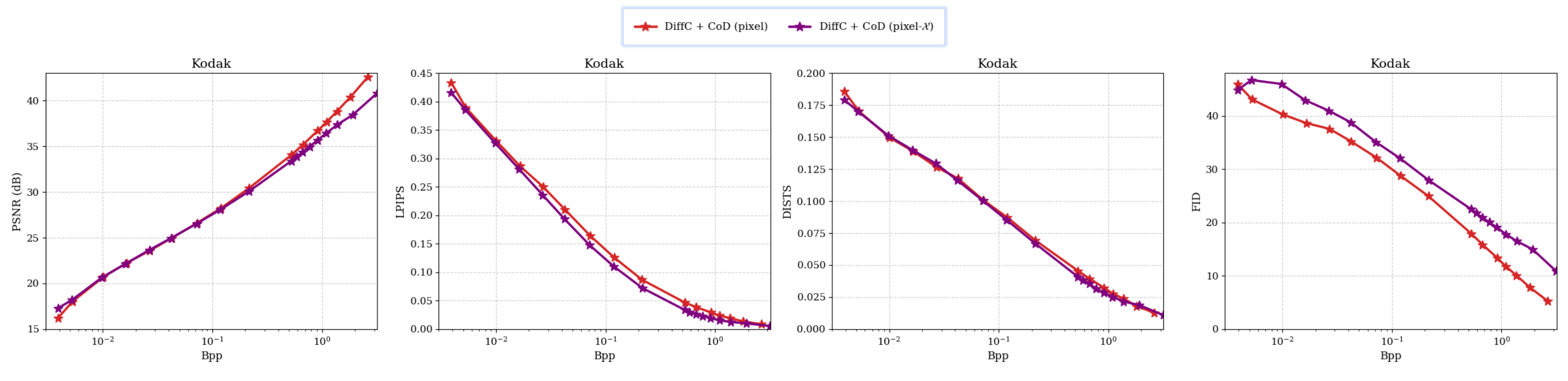}
    \vspace{-7mm}
    \caption{Evaluating $\mathcal{V}$-and $\mathcal{X}$-prediction pixel-space CoD using DiffC on Kodak at $512\times512$.} 
    \vspace{-3mm}
  \label{fig:x_pred_RD}
\end{figure*}

\section{Analysis on Different Downstream Codecs}

In this section, we further demonstrate the capability of CoD as a foundation model through comparing different downstream codecs in Figure~\ref{fig:Compare_downstream_RD}.

\subsection{Comparison of Downstream Codecs}

\noindent\textbf{Zero-shot DiffC}.
No perceptual optimization is involved. It yields the highest PSNR and strong perceptual quality, but has slow encoding speed. Runtime grows with bitrate, and encoding a 4 bpp image can take up to 100 seconds.

\noindent\textbf{Finetuned one-step CoD}.
This scheme attains the best perceptual scores, especially DISTS and FID. The single-step diffusion process enables very fast encoding. However, end-to-end perceptual finetuning with perceptual losses leads to relatively lower PSNR.

\noindent\textbf{CoD-based perceptual optimization for MS-ILLM}.
Although slightly worse than zero-shot DiffC, it requires a lightweight model and offers substantially faster coding, providing a different balance between coding speed and compression ratio.

\subsection{Finetuning Multi-step CoD at Higher Bitrates}

A straightforward downstream compression scheme is to directly finetune multi-step CoD at higher bitrates through diffusion loss. However, this leads to inferior results compared to other schemes. When jointly optimizing a condition under explicit diffusion loss, the optimization becomes biased: the condition tends to maximize the score norm, as shown in~\cite{max_score_norm}. In practice, this drives the decoder toward oversharpened and over-saturated reconstructions, resulting in lower fidelity. This observation is consistent with recent findings that one-step diffusion codecs perform better because they avoid this biased optimization.

The tendency toward a higher score norm also motivates training the foundation model at ultra-low bitrates. Under a strong information bottleneck, the condition can only preserve a small amount of information. The distortion is large at these rates, forcing the available condition to focus on reducing diffusion loss rather than amplifying the score norm. At higher bitrates, reducing diffusion loss becomes easier, leaving room for the condition to optimize for score norm instead. In addition, our unified training strategy and auxiliary losses further help counteract this bias by providing additional compression supervision to the condition network.

\section{Exploring \texorpdfstring{$\mathcal{X}$}{X}-Prediction Pixel Diffusion}
\label{sec:x_pred}

Since $\mathcal{V}$-prediction (velocity prediction) was proposed~\cite{v-pred} and later adopted by Stable Diffusion v2.1, it has become a common practice in latent diffusion models, including the baseline diffusion architectures used in our work~\cite{ddt, pixnerd}. However, recent research by Just Image Diffusion (JiT)~\cite{jit} shows that pixels lie on a low-dimensional manifold, making $\mathcal{X}$-prediction a more effective approach for directly modeling pixel distributions. In this paper, we primarily adopt $\mathcal{V}$-prediction to provide a unified view of latent- and pixel-space compression. Nevertheless, in this section, we conduct experiments to demonstrate that $\mathcal{X}$-prediction can achieve better reconstruction performance.

\begin{table}[h]
\centering
\resizebox{0.9\linewidth}{!}{
    \begin{tabular}{l|cccc}
    \toprule
    Method @ 0.0039 bpp & PSNR & LPIPS & DISTS & FID \\
    \midrule
    CoD (pixel) & 16.21 dB & 0.434 & 0.186 & 46.0\\
    \midrule
    CoD (pixel-$\mathcal{X}$) & \textbf{17.29 dB} & \textbf{0.416} & \textbf{0.179} & \textbf{44.8}\\
    \midrule
    CoD (pixel-$\mathcal{X}$, JiT) & 16.89 dB & 0.422 & 0.183 & 47.4\\
    \bottomrule
    \end{tabular}
}
\caption{Ablation study for $\mathcal{X}$-prediction on Kodak at $512\times512$.}
\label{tab:x_pred}
\end{table}

\noindent\textbf{Comparison on pixel-space CoD}. We follow the JiT training pipeline to build a pixel-space CoD variant using $\mathcal{X}$-prediction and $\mathcal{V}$-loss, denoted as CoD (pixel-$\mathcal{X}$) to distinguish it from our velocity-based pixel-space CoD. To further examine the influence of network design (i.e., the decoupled head~\cite{ddt, pixnerd}) we also substitute our pixel-space diffusion model with JiT’s DiT backbone, referred to as CoD (pixel-$\mathcal{X}$, JiT). As shown in Table~\ref{tab:x_pred}, $\mathcal{X}$-prediction yields significantly better perceptual metrics than $\mathcal{V}$-prediction. In contrast, replacing the decoupled-head design with JiT’s pure DiT structure does not provide performance gains.

\noindent\textbf{Evaluation on downstream DiffC}. In Figure~\ref{fig:x_pred_RD}, we evaluate CoD (pixel-$\mathcal{X}$) on the downstream codec DiffC. Although $\mathcal{X}$-prediction improves performance on the foundation model, it does not consistently benefit DiffC. This is because converting the predicted $x$ into velocity $v$ at time $t$ requires clamping the denominator to be no smaller than 0.05 for stability (following JiT): $v=(x-x_t)\ /\ \textrm{clamp}(1-t, 0.05, 1)$. This clamping leads to inaccurate likelihood estimation near $t=1$. When applied to DiffC, the resulting reconstructions retain slight noisy, i.e., the inversion process does not fully denoise the image, leading to degraded FID. The effect is more pronounced at very high bitrates, where the inaccurate likelihood directly impacts DiffC and causes all metrics to drop relative to $\mathcal{V}$-prediction. These results highlight that different prediction targets offer different strengths, and the choice should be adapted to the task objective.

\begin{table*}[t]
\centering
\resizebox{\linewidth}{!}{
    \begin{tabular}{l|lccccccc}
    \toprule
    Method & \multicolumn{1}{c}{Stage} & Image Resolution & BPP / Total Bits & \#Images & Training Steps & Batch Size & Learning Rate & GPU hours (A100) \\
    \midrule
    CoD & Low-Resolution Pre-Training & \(256 \times 256\) & 0.0156 bpp / 1024 bits & 22.1M & 400 K & \(4 \times 32\) & \(1 \times 10^{-4}\) & $4\times67$ \\
    CoD & High-Resolution Pre-Training & \(512 \times 512\) & 0.0039 bpp / 1024 bits  & 12.8 M & 100 K & \(4 \times 16\) & \(2 \times 10^{-5}\) & $4\times25$ \\
    CoD & Unified Post-Training  & \(512 \times 512\) & 0.0039 bpp / 1024 bits & 12.8 M & 50 K & \(4 \times 16\) & \(2 \times 10^{-5}\) & $4\times24$ \\
    \bottomrule
    \end{tabular}
}
\caption{Detailed configuration of each CoD training stage. The full training process takes 464 A100 GPU hours (approximately 20 days).}
\label{tab:training_detail}
\end{table*}

\begin{table*}[t]
\centering
\resizebox{0.7\linewidth}{!}{
    \begin{threeparttable}
    \begin{tabular}{l|ccc|c|c}
    \toprule
    \multirow{2}{*}{Speed (ms) / Params.} & \multicolumn{3}{c|}{Per-Module Breakdown} & \multirow{2}{*}{Steps} & \multirow{2}{*}{Total}\\
    % \cmidrule(lr){3-4}
    ~ & Conditioner & Diffusion & VAE Decoder & ~ & ~ \\
    \midrule
    Stable Diffusion v1.5 & 203.0 / 3.7 B\tnote{*} & 30.6 / 860 M & 43.4 / 49 M & 25 & 1011 / 4.6 B \\
    \midrule
    Latent-space CoD & 8.3 / 177 M & 21.5 / 676 M  & 43.4 / 49 M & 25 & 589.2 / 901 M \\
    Pixel-space CoD & 8.3 / 177 M & 25.5 / 720 M & - / - & 25 & 645.8 / 897 M \\
    \midrule
    One-Step Pixel-space CoD & 8.3 / 177 M & 16.9 / 513 M\tnote{**} & - / - & 1 & 25.2 / 690 M \\
    \bottomrule
    % \vspace{1mm}
    \end{tabular}
    \begin{tablenotes}
        \footnotesize
        \item[*] Using BLIP2 captioner with at most 32 tokens, as suggested by PerCo (SD)~\cite{perco_sd}.
        \item[**] The AdaLN-Zero layers are omitted since they become redundant when inference operates with a fixed $t=0$.
    \end{tablenotes}
    \end{threeparttable}
}
\caption{Complexity comparison with Stable Diffusion. Average speed (ms) is measured for $512\times512$ on A100.}
\label{tab:complexity}
\end{table*}

\begin{table}[h]
\centering
\resizebox{0.9\linewidth}{!}{
    \begin{tabular}{l|l|ccc}
    \toprule
    ID & Ablation @ 0.0039 bpp & PSNR & LPIPS & FID \\
    \midrule
    A & Flow matching loss & 9.83 dB & 0.576 & 76.8\\
    \midrule
    B & A + Auxiliary Loss   & 15.20 dB & 0.458 & 48.3\\
    C & A + Unified Training & 15.83 dB & 0.433 & 48.4 \\
    \midrule
    D & B + Unified Post-Training & \textbf{16.20 dB} & \textbf{0.433} & \textbf{46.0}\\
    \bottomrule
    \end{tabular}
}
\caption{Ablation study on unified training and auxiliary loss on Kodak at $512\times512$. }
\label{tab:aba_aux}
\end{table}

\section{Additional Details and Results}

This section provides further training details and additional evaluation results, including complexity analysis, additional evaluation metrics and visual comparisons.

\subsection{Training Details}

\subsubsection{Datasets}

We train the model using three publicly available datasets: \textbf{ImageNet-21K}~\cite{imagenet} contains 14.2M images across 21K categories. After removing images with a shorter edge below 256 pixels, 9.3M images remain for $256 \times 256$ pre-training. \textbf{OpenImages}~\cite{openimage} and \textbf{SA-1B}~\cite{sa1b} provide high-resolution images. We use 1.7M directly downloadable images from OpenImages V4 and all 11.1M images from SA-1B for both $256\times256$ and $512\times512$ resolution pre-training. Overall, CoD is trained on 22.1M images. Compared with modern diffusion pipelines, this is a relatively modest data scale, and the only filtering criterion is image resolution rather than extensive data cleaning. We expect that scaling up training with higher-quality data will further improve the performance of CoD.

\subsubsection{Unified Training}

Section~2.2 (in the paper) emphasizes that unified training is crucial for achieving high reconstruction fidelity. Table~\ref{tab:aba_aux} reinforces this observation: unified training (ID = C) provides a clear improvement over the baseline (ID = A). Without unified optimization of the condition and diffusion branches, the condition model tends to encode only structural information while neglecting essential color cues. An intuitive workaround is to impose explicit supervision on the condition learning using an auxiliary loss $\mathcal{L}_{\mathrm{aux}}$.

\noindent\textbf{Auxiliary Loss}.
Given the output of the condition decoder, we attach two lightweight auxiliary prediction heads to (1) reconstruct the input $x$ and (2) predict its DINO~v2~\cite{dino_v2} representation. The auxiliary objective is
\begin{equation}
\mathcal{L}_{\mathrm{aux}} =\mathcal{L}_{\mathrm{aux}}^{\mathrm{MSE}} + 0.5\cdot\mathcal{L}_{\mathrm{aux}}^{\mathrm{DINO}}
\end{equation}
where $\mathcal{L}{\mathrm{aux}}^{\mathrm{MSE}}$ measures pixel reconstruction accuracy, and $\mathcal{L}_{\mathrm{aux}}^{\mathrm{DINO}}$ promotes feature consistency by maximizing the cosine similarity between predicted and ground-truth DINO representations. Each auxiliary head consists of a lightweight three-layer convolutional module, adding negligible training overhead. All auxiliary heads are removed at inference time. As shown in Table~\ref{tab:aba_aux}, auxiliary supervision alone (ID = B) achieves FID competitive with unified training, but exhibits lower PSNR and LPIPS since the diffusion model does not receive direct distortion supervision.

\begin{figure}[t]
  \centering
    \includegraphics[width=\linewidth]{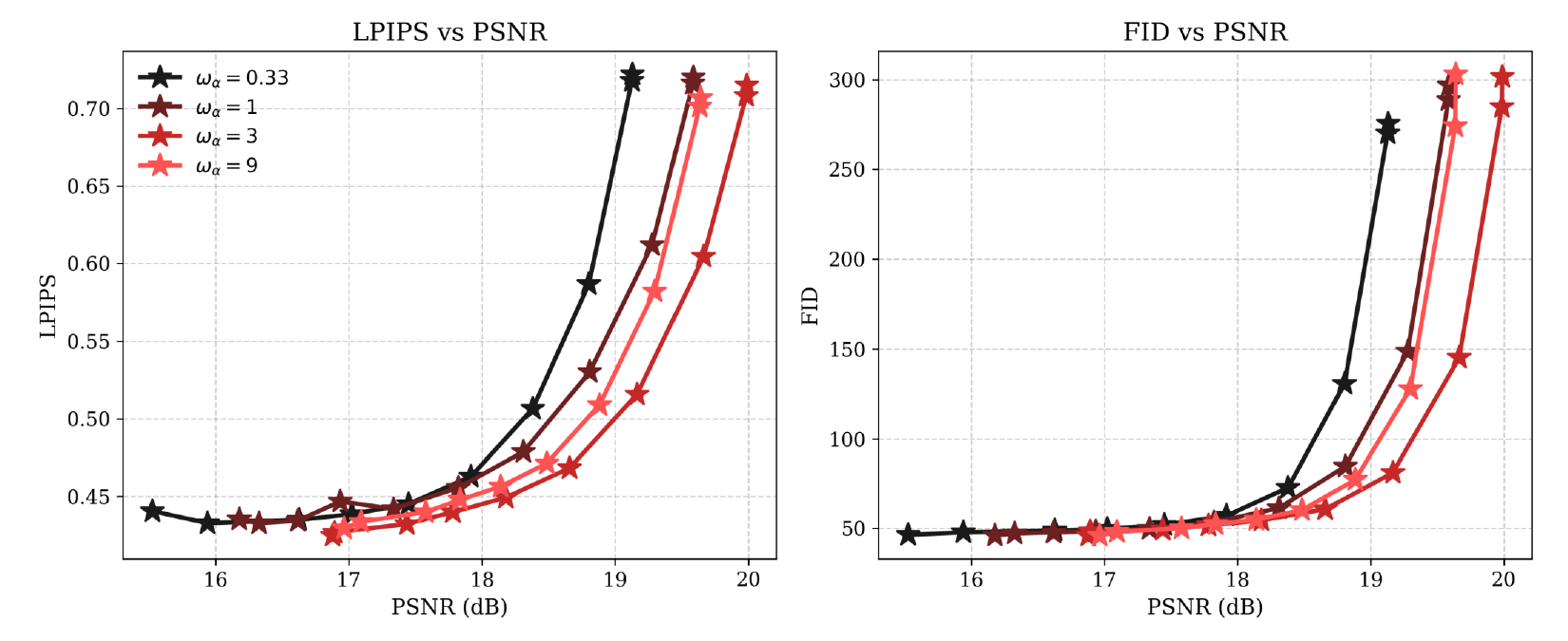}
    \vspace{-7mm}
    \caption{Ablation study on $\omega_\alpha$ in unified post-training. } 
    \vspace{-3mm}
  \label{fig:Ablation_alpha}
\end{figure}

\noindent\textbf{Unified Post-Training}.
Motivated by the complementary strengths of unified training and auxiliary supervision, we combine both approaches to further improve performance. Specifically, we first pre-train CoD with auxiliary loss and then post-train it using unified training. For simplicity, rather than randomly selecting a subset of $\alpha\%$ samples for one-step training, we apply a unified objective to all samples during post-training: $\mathcal{L}_{\text{RF}}=\omega_\alpha \cdot \mathcal{L}_{\text{RF}}^{\text{one-step}}+\mathcal{L}_{\text{RF}}^{\text{multi-step}}$, where $\omega_\alpha$ serves as a weighting factor for the one-step loss component. This brings a little better performance. As shown in Table~\ref{tab:aba_aux}, this unified post-training strategy (ID = D) achieves the best overall results among all configurations.

We adopt $\omega_\alpha = 1$ as the default setting for unified post-training. In Figure~\ref{fig:Ablation_alpha}, we conduct a sensitivity analysis of this parameter, and the results demonstrate that a value of $\omega_\alpha = 3$ yields additional performance gains, suggesting that further optimization of this hyperparameter could enhance overall efficacy.

\begin{table}[t]
\centering
\resizebox{\linewidth}{!}{
    \begin{tabular}{l|ccccccc}
    \toprule
    Metric & PerCo & OSCAR & DiffC & DDCM & OneDC & CoD \\
    \midrule
    User Score (1-5) $\uparrow$ & 3.3 & 2.0 & 1.8 & 1.6 & 3.4 & \textbf{4.2} \\
    \midrule
    CLIP Sim. $\uparrow$ & 0.84 & 0.78 & 0.72 & 0.63 & 0.84 & \textbf{0.89} \\
    Caption Sim. $\uparrow$ & 0.87 & 0.83 & 0.76 & 0.72 & 0.86 & \textbf{0.89} \\
    VQA Acc. $\uparrow$ & 69\% & 66\% & 58\% & 48\% & 71\% & \textbf{80\%} \\
    \bottomrule
    \end{tabular}
}
\caption{Additional evaluation metrics including user study and semantic scores around 0.004 bpp on Kodak.}
\label{tab:additional_eval}
\end{table}

\subsubsection{Multi-Stage Training}

In Table~\ref{tab:training_detail}, we summarize the configuration of all CoD training stages. A key design choice is to keep the total amount of transmitted information fixed across different resolutions. Concretely, we allocate 0.0156 bpp at 0.0156 bpp at $256 \times 256$ and 0.0039 bpp at $512 \times 512$, which correspond to the same total bottleneck size of 1024 bits. The codebook size remains unchanged across resolutions. Instead, we change the downsample ratio in the encoder, i.e., $16\times$ at $256 \times 256$ and $32\times$ at $512 \times 512$.

\subsection{Complexity Analysis}

Table~\ref{tab:complexity} compares the computational cost of Stable Diffusion v1.5 and CoD. Stable Diffusion requires a large captioning model to generate text conditions, whereas CoD relies on a lightweight 177M image encoder and decoder, taking only 8.3 ms to process a $512\times512$ image. The diffusion module in CoD also incurs lower latency and parameter overhead. Pixel-space CoD is slightly slower than latent-space CoD due to the additional decoupled pixel head, but it avoids the expensive VAE decoding, which makes it faster in the few-step regime.

Compared to multi-step diffusion codecs, the one-step pixel-space CoD offers significantly faster inference with a much smaller parameter footprint (see Section~\ref{sec:one-step}). It enables real-time coding, achieving $25.2$ ms per $512\times512$ image. In the future, we hope it can be further accelerated by training a smaller model. Section~\ref{sec:dmd_loss} demonstrates that CoD can be used as a perceptual supervisory signal, suggesting a promising path toward fast inference: \textbf{train a lightweight CoD and distill it into a single step using the large CoD as perceptual supervision}. We leave this direction to future work with the goal of real-time high-resolution diffusion codecs.

\subsection{Additional Evaluation Metrics}

Table \ref{tab:additional_eval} presents additional performance comparison on the Kodak dataset, with all codecs constrained to approximately 0.004 bpp to ensure a fair evaluation. For the subjective user study, 20 participants rated the reconstruction quality on a scale of 1 (lowest) to 5 (highest), from which Mean Opinion Scores (MOS) were derived. To evaluate semantic preservation, we employed three distinct metrics: (1) CLIP-based visual similarity between original and reconstructed embeddings; (2) text-based similarity between BLIP-2 generated captions of the reconstructions and the ground truth; and (3) a multi-choice Visual Question Answering (VQA) benchmark consisting of 10 GPT-4-designed questions per image. Across all subjective and semantic dimensions, CoD consistently outperforms existing codecs, underscoring its superior generative fidelity.

\subsection{Visual Comparison}
\begin{figure*}[t]
  \centering
    \includegraphics[width=\linewidth]{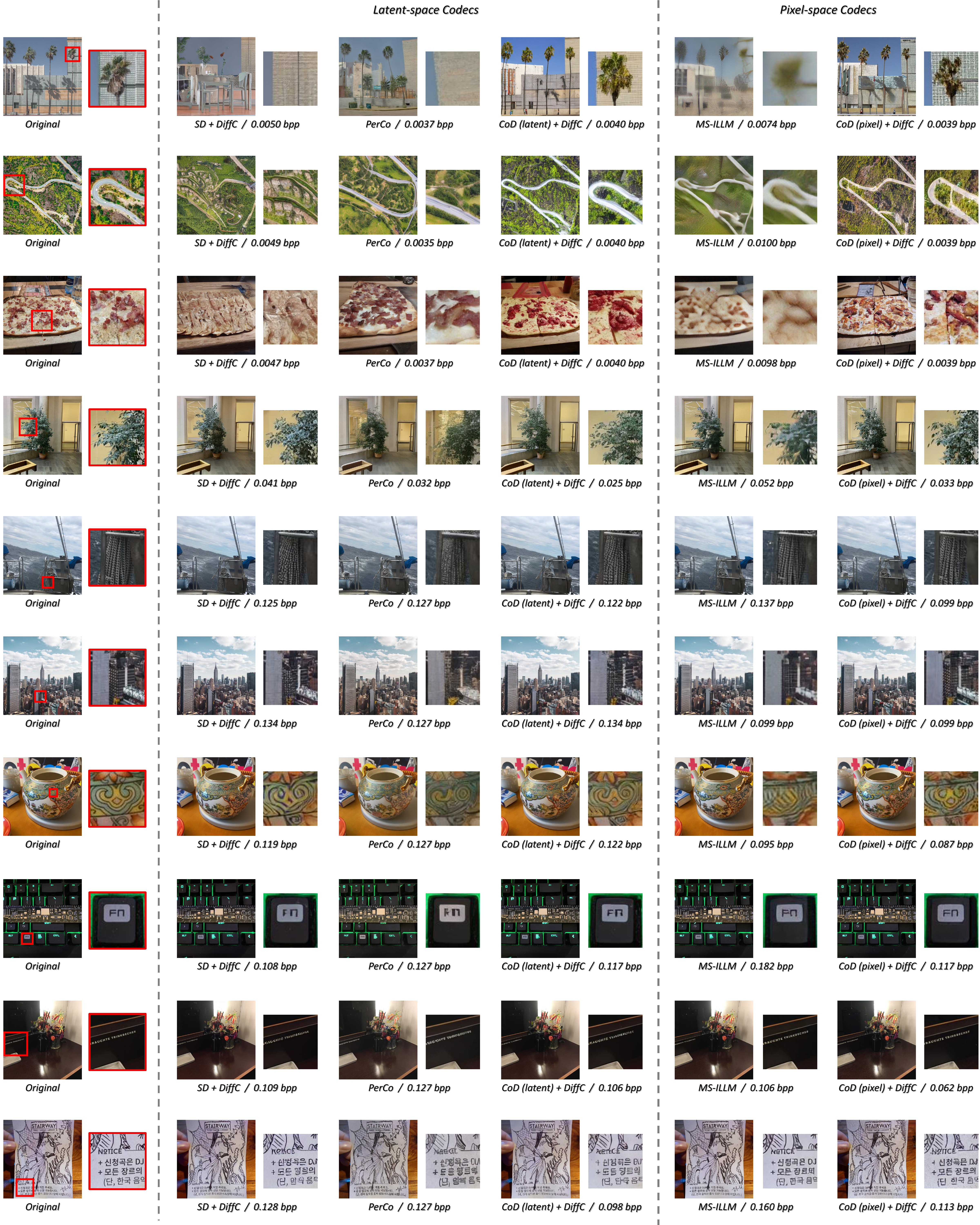}
    \caption{More visualization results on CLIC 2020 test set~\cite{clic}.} 
  \label{fig:Visual_supp}
\end{figure*}

In Figure~\ref{fig:Visual_supp}, we provide more visual comparison examples. Across a wide bitrate range, CoD-based DiffC presents higher perceptual quality than other codecs.

\end{document}